\documentclass[sigconf]{acmart}
\AtBeginDocument{%
  }

\setcopyright{acmlicensed}

\copyrightyear{2026}
\acmYear{2026}
\setcopyright{acmlicensed}
\acmConference[WWW '26] {Proceedings of the ACM Web Conference 2026}{April 13--17, 2026}{Dubai, UAE.}
\acmBooktitle{Proceedings of the ACM Web Conference 2026 (WWW '26), April 13--17, 2026, Dubai, UAE.}
\acmISBN{978-1-4503-XXXX-X/2026/06}
\acmDOI{XXXXXXX.XXXXXXX}

\usepackage{hyperref}       
\usepackage{url}            
\usepackage{booktabs}       
\usepackage{amsfonts}       
\usepackage{nicefrac}       
\usepackage{microtype}      
\usepackage{xcolor}         

\usepackage{graphicx}
\usepackage{subcaption}

\usepackage{amsmath}
\usepackage{mathtools}
\usepackage{amsthm}
\usepackage{algorithm}
\usepackage{algorithmic}
\usepackage{colortbl} 
\usepackage{pifont}

\usepackage{enumitem}
\usepackage{svg}
\usepackage{booktabs}
\usepackage{multirow}
\usepackage{appendix}
\usepackage{wrapfig}

\usepackage[most]{tcolorbox}
\usepackage{dashrule}

\usepackage[capitalize,noabbrev]{cleveref}
\crefname{section}{Sec.}{Secs.}
\Crefname{section}{Section}{Sections}
\Crefname{table}{Table}{Tables}
\crefname{table}{Tab.}{Tabs.}

\newcommand{\ours}[0]{Med-R$^2$\ }

\definecolor{BF6119}{HTML}{BF6119}

\begin{document}

\title{Med-R$^2$: Crafting Trustworthy LLM Physicians via Retrieval and Reasoning of Evidence-Based Medicine}


\author{Keer Lu}

\author{Wentao Zhang}
\authornote{Corresponding Author.}
\email{wentao.zhang@pku.edu.cn}
\affiliation{%
  \department{Center for Data Science, AAIS}
\institution{Peking University}
  \city{Beijing}
  \country{China}
}

\author{Zheng Liang}
\author{Da Pan}
\author{Shusen Zhang}
\author{Guosheng Dong}
\affiliation{%
  \institution{Baichuan Inc.}
  \city{Beijing}
  \country{China}
}

\author{Huang Leng}
\author{Zhonghai Wu}
\author{Bin Cui}
\affiliation{%
  \department{School of Computer Science}
\institution{Peking University}
  \city{Beijing}
  \country{China}
}

\renewcommand{\shortauthors}{Keer Lu et al.}

\begin{abstract}
Large Language Models (LLMs) have exhibited remarkable capabilities in clinical scenarios. 
Despite their potential, existing works face challenges when applying LLMs to medical settings. 
Strategies relying on training with medical datasets are highly cost-intensive and may suffer from outdated training data. 
Leveraging external knowledge bases is a suitable alternative, yet it faces obstacles such as limited retrieval precision and poor effectiveness in answer extraction. 
These issues collectively prevent LLMs from demonstrating the expected level of proficiency in mastering medical expertise. 
To address these challenges, we introduce \textbf{Med-R$^2$}, a novel LLM physician framework that adheres to the \textit{Evidence-Based Medicine (EBM)} process, efficiently integrating retrieval mechanisms as well as the selection and reasoning processes of evidence, thereby enhancing the problem-solving capabilities of LLMs in healthcare scenarios and fostering a trustworthy LLM physician. 
Our comprehensive experiments indicate that \textbf{Med-R$^2$} achieves an improvement of 13.27\% over vanilla RAG methods and even a 4.55\% enhancement compared to fine-tuning strategies, without incurring additional training costs. 
Furthermore, we find that our LLaMA3.1-70B + Med-R$^2$ surpasses frontier models, including GPT-4o, Claude3.5-Sonnet and DeepSeek-V3 by 1.05\%, 6.14\% and 1.91\%. 
Med-R$^2$ effectively enhances the capabilities of LLMs in the medical domain.
\end{abstract}


\received{20 February 2007}
\received[revised]{12 March 2009}
\received[accepted]{5 June 2009}

\maketitle

\setcounter{tocdepth}{0}
\addtocontents{toc}{\protect\setcounter{tocdepth}{0}}

\begin{figure}[ht]
    \begin{center}
    \centerline{\includegraphics[width=\columnwidth]{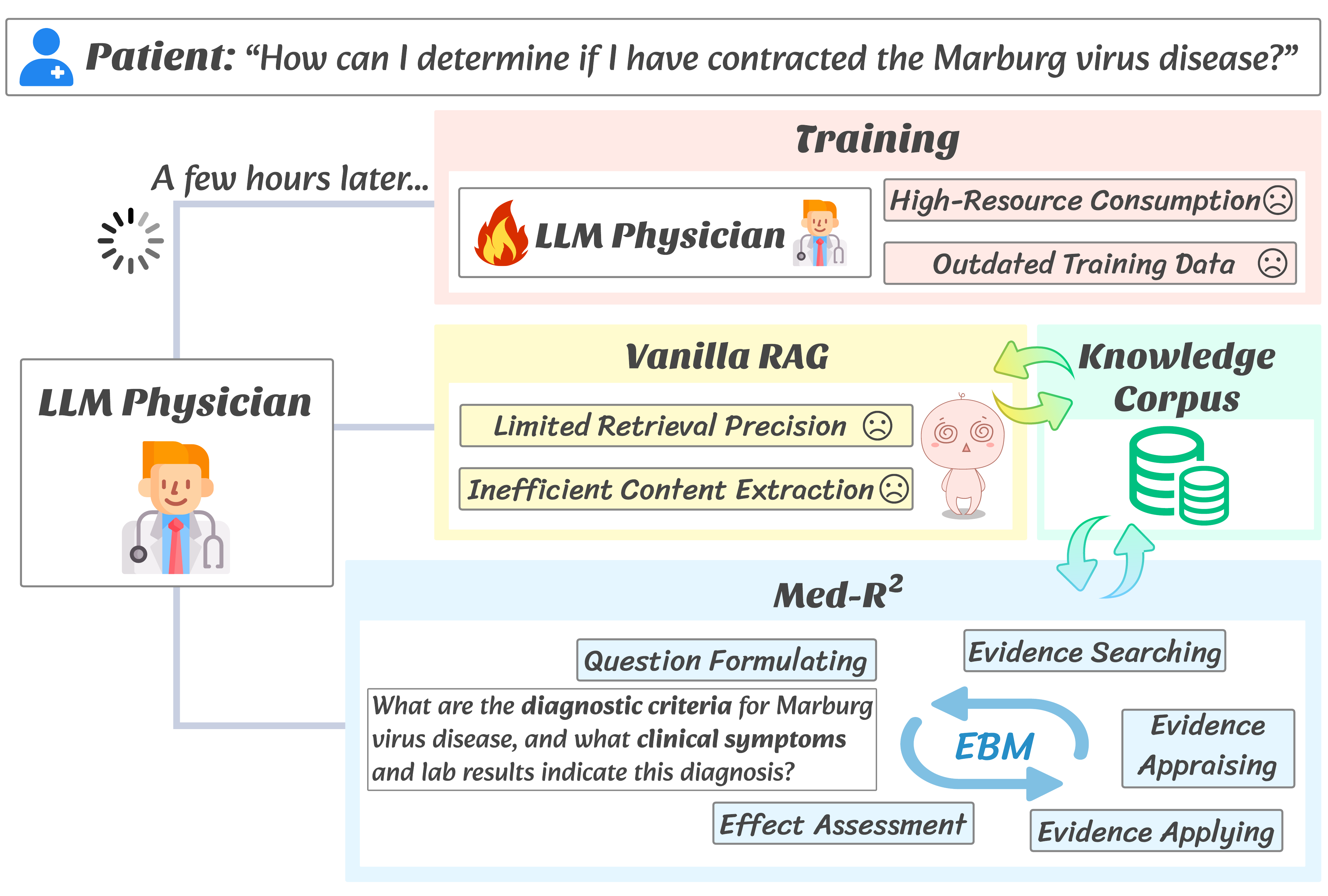}}
    \caption{Comparison of \ours with existing strategies for medical problem-solving.}
    \label{fig:MedRR_intro}
    \end{center}
    \vskip -0.2in
\end{figure}

\section{Introduction}
\label{sec:intro}

Large Language Models (LLMs) have emerged as pivotal tools in the medical domain, redefining the contours of healthcare practice~\cite{abd2023large,yang2023large,thirunavukarasu2023large}. 
Their ability to process and understand vast amounts of unstructured medical data positions them at the forefront of medical research~\cite{clusmann2023future,mumtaz2024llms}, clinical decision-making~\cite{hager2024evaluation,kim2024mdagents}, and patient care~\cite{busch2024systematic,tripathi2024efficient}. 
Despite their remarkable capabilities, LLMs encounter several challenges when applied specifically to healthcare settings:



\textbf{\textit{C1: Inefficiency in Knowledge Acquisition.}} 
Existing approaches predominantly rely on 
training 
LLMs with datasets from the medical domain, the process of which is inherently compute-intensive and resource-demanding, especially as model sizes increase~\cite{singhal2023towards,luo2022biogpt}. 
Moreover, training with outdated datasets can result in a lack of highly specialized expertise, which can lead to suboptimal clinical recommendations or misinform healthcare professionals~\cite{he2022rethinking,kandpal2023large}.

\textbf{\textit{C2: Limited Precision for Medical Retrieval.}} 
Compared to domain-specific training, Retrieval-Augmented Generation (RAG) systems provide a cost-efficient solution, leveraging the external knowledge base to enhance content generation~\cite{lewis2020retrieval}. 
The retrieval quality is critical in RAG systems, where inaccuracy or misinformation can heavily influence the effectiveness of LLMs' augmentation. 
While efforts~\cite{wang2024augmenting,jeong2024improving,long2024bailicai} have been made to improve the retrieval precision, they neglect the specific and highly professional nature of medical knowledge, where \textit{tailored retrieval enhancement} for distinct medical scenarios remains insufficiently explored. 


\textbf{\textit{C3: Low Effectiveness in Answer Extraction.}} 
Considering the constraints imposed by the models' context window length, 
it is essential to critically appraise the retrieved medical evidence for its validity, impact, as well as applicability, and integrate the most pertinent ones with existing clinical expertise for problem-solving~\cite{case1998constructing,guyatt1992evidence}. 
Nonetheless, current studies fail to develop targeted answer extraction methods tailored for healthcare scenarios, where the nuanced evaluation of evidence hierarchies 
is 
required~\cite{sackett1996evidence}. 

To address these challenges, we introduce \textbf{\textit{Med-R$^2$}}, a noval medical LLM framework designed in accordance with the principles of Evidence-Based \textit{\textbf{\underline{Med}}}icine (EBM), conducting outstanding \textit{\textbf{\underline{R}}}etrieval and \textit{\textbf{\underline{R}}}easoning aligned with distinct phases of EBM. 
\textbf{\underline{1)}} For \textbf{\textit{C1}}, we have established a comprehensive external knowledge base to enhance models' medical performances, offering a more cost-effective and flexible alternative to domain-specific training. 
\textbf{\underline{2)}} For \textbf{\textit{C2}}, 
we improve the retrieval precision by refining the original queries according to their respective medical scenarios, while iteratively incorporating chain-of-thought sequences generated from the retrieved content. 
\textbf{\underline{3)}} For \textbf{\textit{C3}}, 
we adopt a coarse-to-fine strategy for document appraising and filtering, 
and select the most pertinent ones supplemented with chain-of-thought demonstrations to assist medical queries addressing. 
Our contributions are as follows: 

\begin{itemize}[leftmargin=*]
    \item \underline{\textit{Challenges in Medical Scenarios.}} 
    Through conducting a quantitative analysis of strategies aimed at enhancing models' medical capabilities, we underscore the challenges prevalent in healthcare scenarios, including high computational consumption as well as poor efficiency in knowledge retrieval and extraction. 
    \item \underline{\textit{LLM Physician Framework.}} 
    We present Med-R$^2$, a novel LLM physician framework that integrates the principles of \textit{evidence-based medicine (EBM)} for clinical problem-solving with outstanding retrieval and reasoning capabilities within medical contexts. 
    \item \underline{\textit{Performance and Effectiveness.}} 
    Extensive experiments indicate that \ours achieves a 14.74\% improvement over the vanilla RAG methods, and even a 3.32\% enhancement compared to the fine-tuning strategies without additional training expenses. Moreover, LLaMA3.1-70B + Med-R$^2$ surpasses frontier models for medical problem-solving, achieving average improvements over GPT-4o, Claude3.5-Sonnet and DeepSeek-V3 by 1.22\%, 5.33\% and 2.80\%. 
\end{itemize}

\section{Related Work and Discussions}
\label{sec:related_work}

\textbf{Evidence-Based Medicine (EBM)} \quad 
EBM refers to the application of the best available research to healthcare, which requires evidence integration with clinical expertise and patient values~\cite{sackett1996evidence,guyatt1992evidence,sackett1997evidence}. 
Clinical questions can be categorized into several types, including \textit{diagnosis, therapy, prognosis, etiology, prevention, cost,} etc.~\cite{case1998constructing}. 
Each category intersects with EBM principles by emphasizing the 
collection, evaluation, and application of the best retrieved evidence to inform medical decision-making~\cite{guyatt2000users}, as detailed in \Cref{sec:EBM}.

\begin{figure*}[ht]
    \begin{center}
    \centerline{\includegraphics[width=\linewidth]{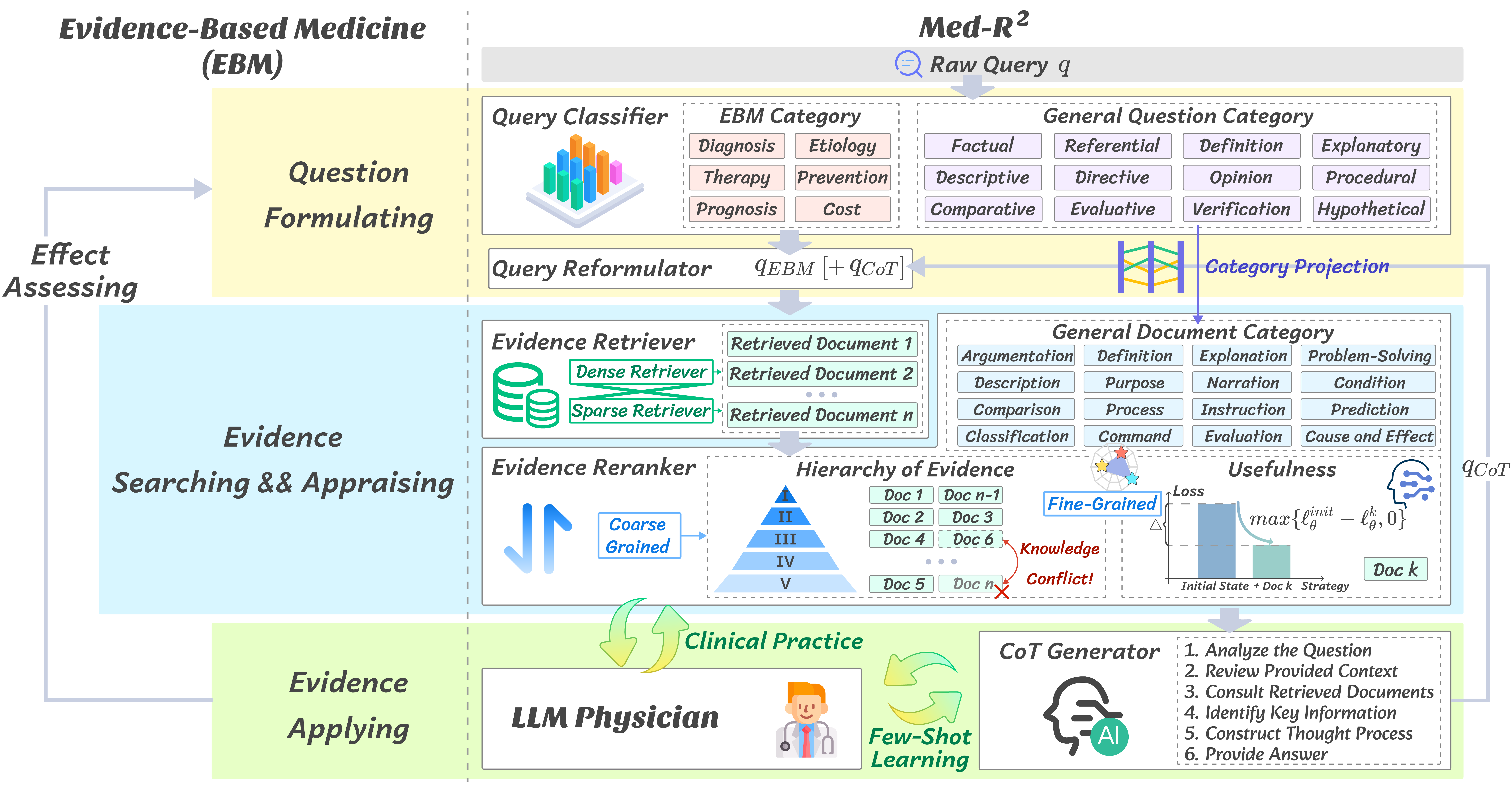}}
    \caption{An illustration of Med-R$^2$'s process, adhering to the Evidence-Based Medicine (EBM) workflow. We first categorize the query by EBM and general question types. Queries are then reformulated according to established EBM classification templates to ensure precision and relevance. In the evidence searching and appraising stages, we employ a coarse-to-fine strategy to retrieve, filter, and re-rank the evidence documents within the knowledge base. CoT sequences are then generated from processed evidence to refine retrieval space, iterating to ensure the robustness.
    }
    \label{fig:pipeline}
    \end{center}
    \vskip -0.2in
\end{figure*}

\textbf{LLMs for Medical Domain} \quad 
As the application of LLMs expands, their deployment in the medical domain has become a widely discussed topic~\cite{clusmann2023future,zeng2020meddialog}. 
Recent studies have concentrated on the direct use of 
medical data for the pretraining or fine-tuning of LLMs~\cite{thirunavukarasu2023large,singhal2023large}. 
Prominent open-source milestones include ChatDoctor~\cite{li2023chatdoctor} which integrates real-world doctor-patient communication data for training, PMC-LLaMA~\cite{wu2024pmc} pretrained on 4.9 million medical literature records, and MEDITRON~\cite{chen2023meditron}, a scaling series of medical pretrained models.
However, such extensive training can be computationally intensive. 
In contrast, Retrieval-Augmented Generation (RAG) systems offer a more efficient alternative, achieving comparable results with reduced training costs and enhancing the model's precision in locating and leveraging knowledge.

\textbf{Retrieval-Augmented Generation (RAG)} \quad 
The concept of RAG~\cite{lewis2020retrieval} was introduced as a powerful framework for integrating external knowledge into natural language generation tasks, enhancing the accuracy and relevance of generated outputs across various domains~\cite{gao2023retrieval,zhao2024retrieval,izacard2023atlas}.
In the medical field, RAG has been widely used to improve LLMs' analytical performances by utilizing external medical knowledge from sources such as medical papers, textbooks, guidelines, and entries~\cite{jin2023retrieve,zakka2024almanac,xiong2024benchmarking,xiong2024improving}. 
However, while there have been efforts dedicated to optimizing the individual components of RAG pipelines~\cite{wang2024augmenting,jeong2024improving,asai2023self,jeong2024adaptive}, research that integrates the unique characteristics and requirements of the medical domain remains in its infancy.
In this study, we incorporate the principles of \textit{Evidence-Based Medicine} (\textit{EBM}) into medical RAG systems to better address the special demands of healthcare.

\section{Med-R$^2$}
\label{sec:method}

In this section, we discuss our \ours framework, illustrated in \Cref{fig:pipeline}. 
\ours is designed around the Evidence-Based Medicine (EBM) workflow, encompassing the stages of clinical question formulation~(\Cref{subsec:query_reformulation}), evidence retrieval and appraisal~(\Cref{subsec:evidence_searching_appraising}), evidence application~(\Cref{subsec:evidence_applying}), and effect assessment~(\Cref{subsec:effect_assessment}).

\subsection{Question Formulation}
\label{subsec:query_reformulation}



\begin{figure}[ht]
    \centering
    \includegraphics[width=.45\textwidth]{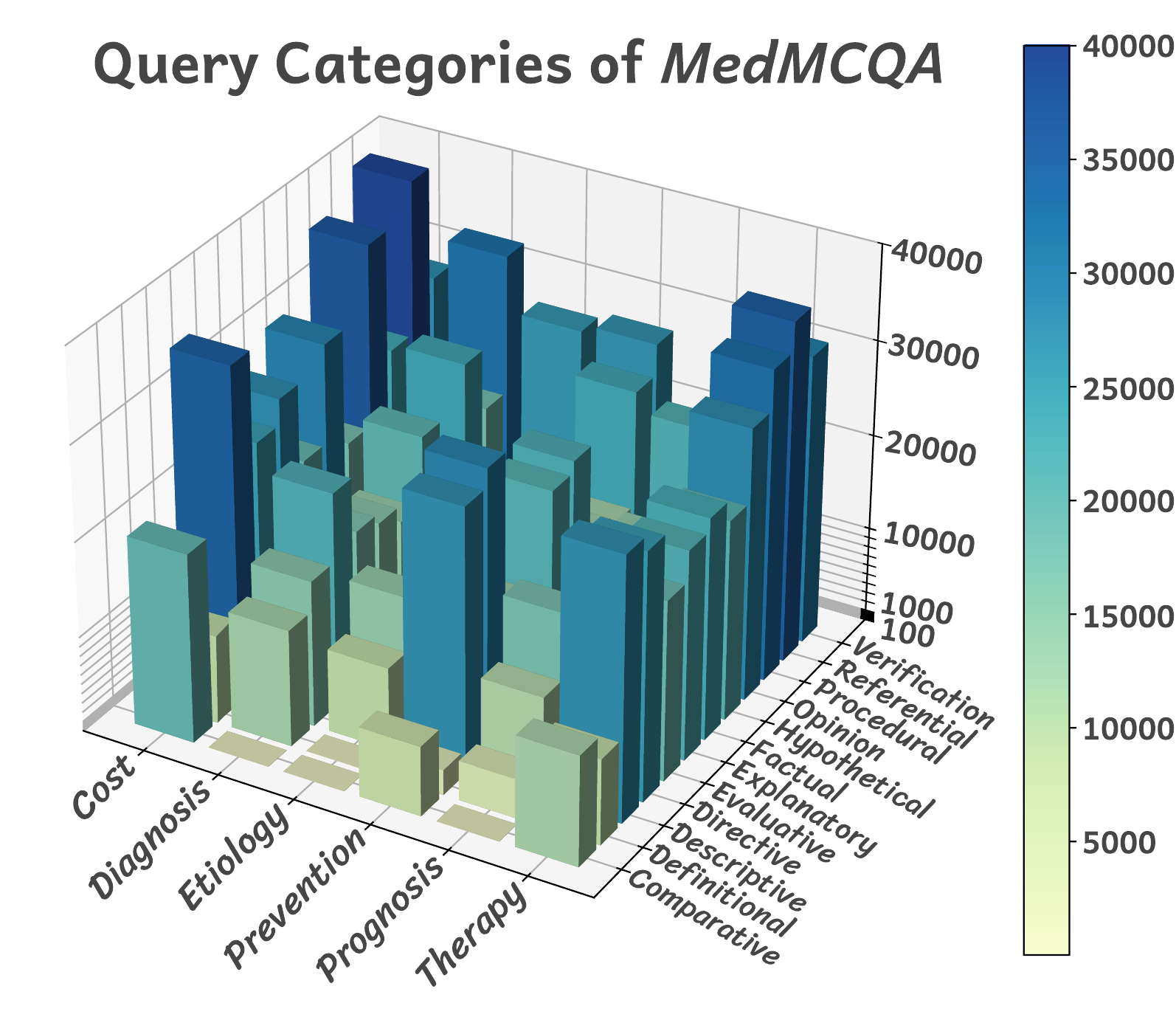}
    \caption{Query category of \textbf{\textit{MedMCQA}}. We employ a logarithmic scale (base 10) on the z-axis, ranging from 1 to 40000, to represent the wide range of values.}
    \label{fig:MedMCQA_query_category}
    \vskip -0.2in
\end{figure}

In the medical domain, the efficacy of information retrieval is closely tied to the professionalism of the query. A well-crafted, professional query that includes precise descriptions of medical symptoms can markedly enhance the accuracy and relevance of the documents retrieved from knowledge bases. Conversely, non-standard terms and isolated numerical values frequently hinder effective information retrieval. 
Moreover, the focus of the desired response varies depending on the type of clinical consultation. 
For instance, in queries pertaining to \textit{etiology}, users seek insights into potential causes of a condition, including risk factors, pathogens, or genetic predispositions. In contrast, for \textit{prognosis}-related queries, users are interested in understanding the long-term outcomes or patient prognoses, such as survival rates or recurrence probabilities.

The clinical question formulation stage consists of two components: \textbf{query classifier} and \textbf{query reformulator}. 
The query classification encompasses two dimensions: \textit{Evidence-Based Medicine (EBM) categories} and \textit{general natural language question types}. 
Specifically, we have delineated six distinct EBM categories and twelve general question categories, as illustrated in \Cref{fig:pipeline}. 
The application of this classification scheme to the MedMCQA dataset yielded the categorized results presented in \Cref{fig:MedMCQA_query_category}. 
For the classification task, we have manually annotated 100 samples to fine-tune Qwen2.5-72B-Instruct\footnote{https://huggingface.co/Qwen/Qwen2.5-72B-Instruct}, and employed the trained model as our classifier.
We perform domain-specific reformulations of the original queries based on their respective classes of the EBM categories to align with the professional context. 
Meanwhile, the general question categories are utilized as one of the criteria to rerank retrieved documents, thereby prioritizing those that best match the current query's intent and document type preferences described in \Cref{fig:query_doc_projection}. 
Further details 
can be found in \Cref{sec:appendix_question_formulation}.


\subsection{Evidence Searching \&\& Appraising}
\label{subsec:evidence_searching_appraising}

The retrieval and appraisal of evidence constitute one of the most critical stages in EBM. 
The professional reformulating of queries in the preceding stages~(\Cref{subsec:query_reformulation}) aims to enhance the precision of retrieving relevant evidence documents. 
This stage comprises two key components: \textbf{evidence retriever} and \textbf{evidence reranker}.

\subsubsection{Evidence Retriever}
\label{subsubsec:retriever}

To construct a more comprehensive medical knowledge base adaptable to diverse healthcare scenarios, we have amassed a collection of medical data to build our \textbf{knowledge corpus}. 
The final medical knowledge base employed for retrieval comprises four distinct types of resources: \textit{academic papers}, \textit{entries}, \textit{books}, and \textit{guidelines}, with details of the data sources and statistics depicted in \Cref{tab:medical_knowledge_corpus}. 
For resources with extensive content such as academic papers, books, and guidelines, we first perform content segmentation with a threshold set at 10,000 tokens. 
We prioritize dividing the content based on natural chapters. If natural chapters cannot be identified or exceed the threshold, we resort to truncation according to the predefined limit.


\begin{table}[t]
    \centering
    \caption{\label{tab:medical_knowledge_corpus}
            Overall statistics of medical knowledge resources.}
    \resizebox{\columnwidth}{!}{
        \begin{tabular}{cccccc}
            \toprule
            \multirow{2}{*}{Source Type} & \multirow{2}{*}{\#Volume} & \multicolumn{4}{c}{\#Tokens / Doc}  \\
             \cline{3-6}
             & & Max & Min & Mean & Medium \\
            \midrule
                Academic Papers &  600,000 & 10,643 & 279 & 3,820 & 3,097 \\
                Entries & 470,000 & 6,538 & 56 & 1,387 & 1,962 \\
                Books & 10,000  & 15,384 & 524 & 4,083 & 4,319 \\
                Guidelines & 10,000  & 4,981 & 74 & 1,100 & 1,778 \\
            \midrule
                Total & 1,090,000 & 15,384 & 56 & 2,748 & 2,219 \\
            \bottomrule
        \end{tabular}
    }
\end{table}


We integrate multiple types of retrievers to optimize our retrieval performance, incluing BGE-Large-EN-v1.5\footnote{https://huggingface.co/BAAI/bge-large-en-v1.5} for dense retrieval and SPLADE-v3\footnote{https://huggingface.co/naver/splade-v3} for sparse retrieval. 
We then consolidate the documents retrieved through both methods into a unified collection that contains $n$ documents, $\mathcal{D} = \left \{ d_i \right \}_{i=1}^n = \mathcal{D}^D \cup \mathcal{D}^S$, where $\mathcal{D}^D$ comprises documents obtained by dense retrieval, and $\mathcal{D}^S$ represents those acquired via sparse retrieval.

\subsubsection{Evidence Reranker}
\label{subsubsec:evidence_reranker}

For the document collections retrieved in \Cref{subsubsec:retriever}, we employ a \textbf{\textit{coarse-to-fine}} strategy for reranking. 
we initially utilize the BGE-Reranker-v2-M3\footnote{https://huggingface.co/BAAI/bge-reranker-v2-m3} to rerank $\mathcal{D} = \left \{ d_i \right \}_{i=1}^n$ based on their semantic relevance to the current query at a \textit{coarse} granularity, returning the top $k$ documents\footnote{During the previous retrieval phase, documents of all retrieval types are subjected to a \textit{unified ranking}, within which the top-$k$ documents are selected for further processing.}, 
denoted as $\mathcal{S}^c = \left \{ d_i \right \}_{i=1}^k$. At this stage, the total length of these $k$ documents significantly exceeds the context limit of the model. 
Then we conduct a \textit{fine}-grained reranking that integrates three distinct criteria for the current $k$ documents to get $\mathcal{S}^f = \left \{ d_i \right \}_{i=1}^{k'}$, enabling the model to provide answers based on the most effective retrieved documents within the limited length of the context window. 
The fine-grained reranking score can be formulated as: 
\begin{gather}  
\mathcal{F}\left( x\right) = f_h\left( x\right ) \cdot f_g\left( x\right ) \left(1 + \alpha \cdot f_u\left( x\right ) \right)
\label{eq:fine_grained_reranking}    
\end{gather}
where $f_h\left( x\right )$, $f_u\left( x\right )$ and $f_g\left( x\right )$ refer to scores of each refined evaluation criteria, while $\alpha$ is the non-negative hyper-parameter for weight controlling. 
The detailed implementation and explanation for the derivation of \Cref{eq:fine_grained_reranking} can be found in \Cref{sec:appendix_evidence_reranking}.

\begin{figure}[t]
    \begin{center}
    \centerline{\includegraphics[width=\linewidth]{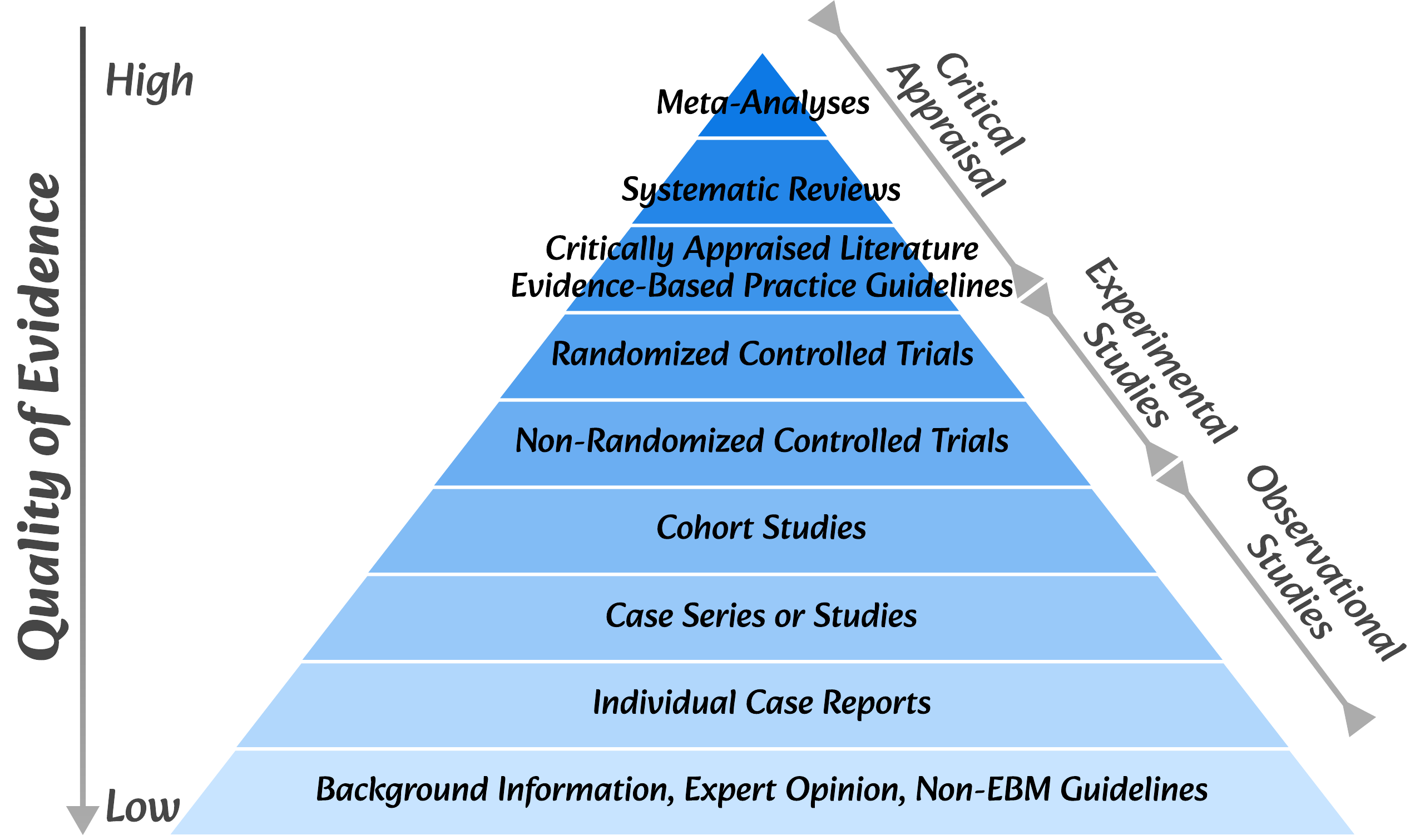}}
    \caption{
    \textbf{\textit{Hierarchy of evidence}}. 
    The base of the pyramid represents the lowest quality, while the apex denotes the highest.
    }
    \label{fig:hierarchy_of_evidence}
    \end{center}
    \vskip -0.2in
\end{figure}

\textbf{Hierarchy of Evidence} \quad 
Medical knowledge encompasses 
facts and theories that are not always consistent, and sometimes even contradictory. 
This criterion serves dual purposes: \textit{scoring} and \textit{conflict filtering}. 
The recalled documents that have undergone coarse-grained reranking are first categorized according to their evidence levels. 
Subsequently, contents of these documents are analyzed for conflicting facts, where the documents that contain conflicting facts and have lower evidence ratings are filtered out. 
Formally, each retrieved document $d_i \in \mathcal{S}^c$ is associated with an integer evidence level $e$, where $e \in \left \{ x \in \mathbb{Z} \mid 1 \leq x \leq 9 \right \}$, with 1 indicating the highest level of credibility, as depicted in \Cref{fig:hierarchy_of_evidence}:
\begin{gather}  
f_h\left( x\right ) = 9 - \left( e_x - 1\right )
\label{eq:f_h}    
\end{gather}
By applying the evidence assessment criteria, the retrieved documents are categorized into multiple evidence levels, and professional analysis of evidence at different levels of authority is performed, preventing misjudgments caused by information clutter.

\begin{figure}[ht]
    \vskip 0.2in
    \begin{center}
    \centerline{\includegraphics[width=\linewidth]{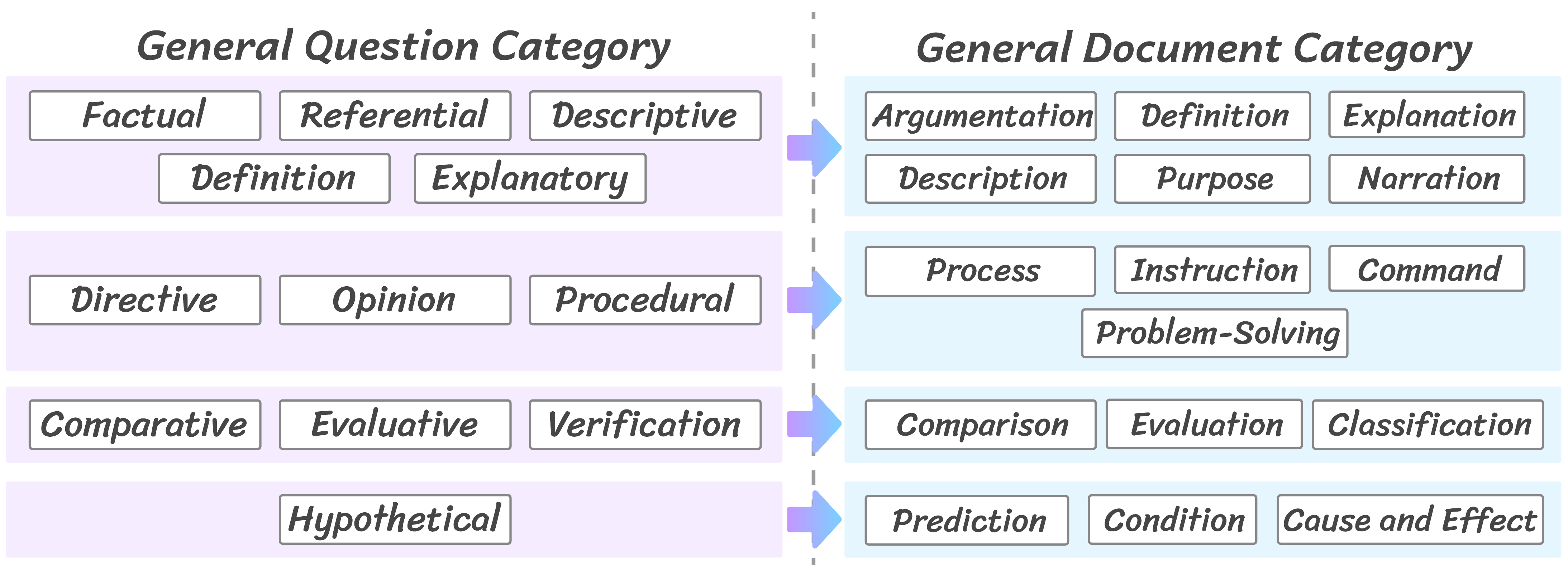}}
    \caption{
    \label{fig:query_doc_projection}
    Query Document Projection.}
    \end{center}
    \vskip -0.2in
\end{figure}

\textbf{Usefulness} \quad 
The usefulness ranker is employed to assess the contribution of retrieved documents to the answering process. Specifically, we quantify the usefulness of a document by measuring the difference in loss before and after using the retrieved document to answer the question. This is achieved with a lightweight proxy model that evaluates the impact of the document on the answer's quality, which can be written as: 
\begin{gather}  
f_u\left( x\right ) = \max \left \{ \ell_\theta^{init} - \ell_\theta^x, 0 \right \}
\label{eq:f_u}
\end{gather}
where $\ell_\theta^{init}$ indicates the loss without referring to any retrieved documents, while $\ell_\theta^x$ represents that informed by document $x$.

\textbf{General Document Category} \quad 
It corresponds to the mapping of \textit{general natural language question} types discussed in \Cref{subsec:query_reformulation}. 
Different categories of questions desire distinct answer structures. 
For instance, questions regarding procedural steps are ideally answered by documents that describe processes rather than those that define concepts. 
To address this, we categorize the retrieved documents into 16 document types, denoted as $C$, as outlined in \Cref{tab:document_category}, and score them based on their alignment with the response type preferred by the original query, as depicted in \Cref{fig:query_doc_projection}. 
The scoring function is defined as:
\begin{equation}
f_g\left( x\right ) = \sum_{j=1}^{|C^e|} p(x | c_j) 
\end{equation}
where $C^e = \left \{ c_j \right \}_{j=1}^{|C^e|}$ stands for the list of expected document types ($C^e \subset C$), and $p(x | c_j)$ represents the probability that the document $x$ belongs to the current expected type $c_j$ from $C^e$. 
Details of computational procedures are provided in \Cref{alg:document_category_rerank}.

\begin{algorithm}[ht]
\caption{Scoring Based on General Document Category Classification}
\label{alg:document_category_rerank}
\small
\begin{flushleft}
\textbf{Input}: Coarsely reranked documents $\mathcal{S}^c$, Query-document mapping $\mathcal{A}$ based on \Cref{fig:query_doc_projection}, classifier $M_P$, List of general document categories $C$ \\ 
\textbf{Parameter}: List of current expected document categories $C^e$, $C^e \subset C$ \\
\mbox{\textbf{Output}: Scores based on document category $f_g(S^c)$} \\
Define $q^c$: category of the query used for retrieving \\
Define $\vec{p}$: category probability distribution of document $d$ \\
\end{flushleft}
\begin{algorithmic}[1]
\FOR{each document $d_i$ in $\mathcal{S}^c$}
    \STATE /* \textit{Step 1: Expected Category Mapping} */
    \STATE Obtain expected categories $C^e = \mathcal{A}(q_{di}^c)$, where $C^e \subset C$ 
    \STATE /* \textit{Step 2: Document Category Probability Inference} */
    \STATE Provide category probability distribution of $d_i$ referring to $M_P$: \\  $$\vec{p_i} = \{p(d_i | c_j)\}_{j=1}^{|C|} \gets M_P(d_i)$$
    \STATE /* \textit{Step 3: Statistics Aggregation} */
    \STATE Calculate the sum of probabilities across all expected categories: \\ \centering{$f_g\left( d_i\right ) = \sum_{j=1}^{|C^e|} p(d_i | c_j) $, where $c_j \in C^e$}
\ENDFOR
\STATE \textbf{Return} $f_g(S^c) = \left \{ f_g(d_i) \right \}_{i=1}^{|S^c|}$
\end{algorithmic}
\end{algorithm}

\subsection{Evidence Applying}
\label{subsec:evidence_applying}

Through the comprehensive evaluation and reranking of retrieved documentary evidence in \Cref{subsubsec:evidence_reranker}, we aim to ensure the application of the highest-quality available evidence to decision-making in the medical field. 
This process extends beyond merely applying research findings, where integrating the professional reasoning and judgment is essential. 
Therefore, it is imperative to make \textit{professional and reasoned inferences} based on the retrieved documents.

\textbf{CoT Generator} \quad 
This module constructs a chain-of-thought reasoning process based on the original medical query and the retrieved evidence documents\footnote{When ablating the integration of CoT sequences at the onset in \Cref{sec:ablation}, the CoT sequences are curated solely from the original medical query during \textit{the initial iteration}, \textit{excluding} the retrieved evidence documents.}. 
It serves dual functions: (1) \textit{Component of Query Reformulation}: It contributes to the subsequent query reformulation process, facilitating the retrieval of evidence documents relevant to the question. 
In \Cref{sec:ablation}, we also clarify the rationale behind \underline{not} incorporating models' generated CoT sequences into the query reformulator for medical knowledge retrieval \underline{at the initial iteration}.
(2) \textit{Few-Shot Learning Instance}: It provides few-shot examples for the LLM physician (target model) intended for downstream task evaluating, demonstrating how to analyze the retrieved evidence, identify the key information, and address medical queries effectively.

\subsection{Effect Assessment}
\label{subsec:effect_assessment}


Our assessment of evidence is centered on the following aspect: the \textit{stability} of evidence document retrieval across different stages. 
\Cref{alg:evidence_assessment} delineates the evidence evaluation process and the iterative retrieval loop for the query corpus within the pipeline, where we determine the number of iterations for optimization termination based on our assessment of such factor. 
This ensures that when evaluating the LLM physician (target model) on medical tasks, the retrieved documents provided are effective and robust.

\begin{algorithm}[ht]
\caption{Evidence Assessment and Iterative Retrieval Loop for Query Corpus}
\label{alg:evidence_assessment}
\small
\begin{flushleft}
\textbf{Input}: Query corpus $\mathcal{Q}$, CoT generator $M_{CoT}$, Context window length $w$, Hyperparameters: alteration threshold $\delta$, maximum iterations $T$ \\ 
\textbf{Params}: Coarsely reranked documents $\mathcal{S}^c$, finely reranked documents $\mathcal{S}^f$ \\
\mbox{\textbf{Output}: Selected evidence documents $\mathcal{S}_{\mathcal{Q}-top}^{f}$ and CoT sequences $\mathcal{Q}^{CoT}$} \\
Define $q^{CoT}$: chain-of-thought sequence generated based on query and associated evidence documents \\
Define $E(d)$: embedding of document $d$ \\
\end{flushleft}
\begin{algorithmic}[1]
\STATE $\mathcal{S}_{\mathcal{Q}-top}^{f}, \mathcal{Q}^{CoT} \gets [\,],\, [\,]$
\FOR{each query $q_i$ in $\mathcal{Q}$}
    \FOR{$t = 1, 2, \ldots, T$}
        \STATE Reformulate query $q_i$ (\Cref{subsec:query_reformulation}), search and appraise the retrieved evidence (\Cref{subsec:evidence_searching_appraising})
        \STATE /* \textit{Step 1: CoT Generation} */
        \STATE Select top $k$ documents from $\mathcal{S}_i^{f^{(t)}}$ referring to $w$: $$\mathcal{S}_{i-top}^{f^{(t)}} \leftarrow \text{SelectTopK}(\mathcal{S}_i^{f^{(t)}}, k, w)$$
        \STATE Generate chain-of-thought sequence based on $q_i$ and $\mathcal{S}_{i-top}^{f^{(t)}}$: $$q^{CoT^{(t)}}_i \leftarrow M_{CoT}(q_i, \mathcal{S}_{i-top}^{f^{(t)}})$$
        \STATE /* \textit{Step 2: Evidence Assessment} */
         \STATE Compute semantic stability of docs in consecutive iterations: $$\vec{\mu}^{(t)}_i \leftarrow ||\frac{1}{|\mathcal{S}_i^{c^{(t)}}|} \sum_{d_i \in \mathcal{S}_i^{c^{(t)}}} E(d_i) - \frac{1}{|\mathcal{S}_i^{c^{(t-1)}}|} \sum_{d_i \in \mathcal{S}_i^{c^{(t-1)}}} E(d_i)||$$
        \STATE /* \textit{Step 3: Termination Condition Evaluation} */
        \IF{$\vec{\mu}^{(t)}_i < \delta$}{
            \STATE $\mathcal{S}_{\mathcal{Q}-top}^{f}.append(\mathcal{S}_{i-top}^{f^{(t)}})$, $\mathcal{Q}^{CoT}.append(q^{CoT^{(t)}}_i)$
            \STATE \textbf{Break}
        }
        \ENDIF
    \ENDFOR
    \STATE $\mathcal{S}_{\mathcal{Q}-top}^{f}.append(\mathcal{S}_{i-top}^{f^{(T)}})$, $\mathcal{Q}^{CoT}.append(q^{CoT^{(T)}}_i)$
\ENDFOR
\STATE \textbf{Return} Selected documents $\mathcal{S}_{\mathcal{Q}-top}^{f}$ and CoT sequences $\mathcal{Q}^{CoT}$
\end{algorithmic}
\end{algorithm}

\section{Experiments and Results}
\label{sec:experiment}

\begin{table*}[tb]
    \centering
    \setlength{\tabcolsep}{1.4pt}
    \caption{\label{tab:main_results} 
    Comparison of \ours with baselines. The best and second best are in \textbf{bold} and \underline{underlined}. 
} 
    \resizebox{\textwidth}{!}{
        \begin{tabular}{ccccccccccc|c}
            \toprule
                \multirow{2}{*}{Model} & \multirow{2}{*}{Method} & \multicolumn{1}{|c}{MedQA-USMLE} & MedQA-MCMLE & MedMCQA &  \multicolumn{1}{|c}{PubMedQA} & MMLU-Med & NEJMQA & MedXpertQA & RareArena-RDC & RareArena-RDS & \multirow{2}{*}{Avg.} \\
            \cline{3-11}
             & & \multicolumn{3}{|c}{\textit{Within-Dataset Fine-Tuning}} &  \multicolumn{6}{|c|}{\textit{Cross-Dataset Fine-Tuning}} &  \\
            \midrule
            \multicolumn{12}{c}{\cellcolor[HTML]{EFEFEF}\textbf{Frontier Models}} \\
            \midrule
            GPT-4o & -- & \multicolumn{1}{|c}{84.95} & 79.28 & 74.60 & \multicolumn{1}{|c}{77.20} & 84.44 & 72.20 & 30.37 & 73.76 & 47.50 & \multicolumn{1}{|c}{\cellcolor[HTML]{FFFBCF}69.31} \\
            Claude3.5-Sonnet & -- & \multicolumn{1}{|c}{83.20} & 72.54 & 68.80 & \multicolumn{1}{|c}{76.40} & 83.90 & 68.52 & 21.31 & 68.46 & 50.78 & \multicolumn{1}{|c}{\cellcolor[HTML]{FFFBCF}65.99} \\
            DeepSeek-V3 & -- & \multicolumn{1}{|c}{80.93} & 77.82 & 74.30 & \multicolumn{1}{|c}{73.60} & 87.67 & 74.23 & 24.16 & 75.09 & 45.72 & \multicolumn{1}{|c}{\cellcolor[HTML]{FFFBCF}68.73} \\
            \midrule
            \multicolumn{12}{c}{\cellcolor[HTML]{EFEFEF}\textbf{Open-Sourced Medical Models}} \\
            \midrule
            PMC-LLaMA-7B & -- & \multicolumn{1}{|c}{38.65} & 35.30 & 32.40 & \multicolumn{1}{|c}{52.26} & 26.87 & 35.63 & 11.02 & 28.96 & 21.67 & \multicolumn{1}{|c}{\cellcolor[HTML]{FFFBCF}31.42} \\
            MEDITRON-7B & -- & \multicolumn{1}{|c}{46.76} & 42.78 & 36.83 & \multicolumn{1}{|c}{58.32} & 43.60 & 38.80 & 11.74 & 35.68 & 26.50 & \multicolumn{1}{|c}{\cellcolor[HTML]{FFFBCF}37.89} \\
            PMC-LLaMA-13B & -- & \multicolumn{1}{|c}{40.38} & 33.07 & 37.55 & \multicolumn{1}{|c}{54.63} & 46.85 & 40.60 & 11.44 & 39.86 & 28.98 & \multicolumn{1}{|c}{\cellcolor[HTML]{FFFBCF}37.04} \\
            MEDITRON-70B & -- & \multicolumn{1}{|c}{55.05} & 52.21 & 53.50 & \multicolumn{1}{|c}{72.84} & 71.45 & 66.97 & 18.05 & 74.81 & 46.58 & \multicolumn{1}{|c}{\cellcolor[HTML]{FFFBCF}56.83} \\
            \midrule
            \multicolumn{12}{c}{\cellcolor[HTML]{EFEFEF}\textbf{Open-Sourced Base Models}} \\
            \midrule
            \multirow{5}{*}{LLaMA3.1-8B} & Direct Response & \multicolumn{1}{|c}{31.16} & 41.45 & 30.02 & \multicolumn{1}{|c}{36.17} & 37.12 & 50.41 & 14.90 & 41.85 & 22.16 & \multicolumn{1}{|c}{\cellcolor[HTML]{FFFBCF}33.91} \\
             & Vanilla RAG & \multicolumn{1}{|c}{55.80} & 59.38 & 35.91 & \multicolumn{1}{|c}{\underline{47.10}} & 43.54 & 52.01 & 15.50 & 56.23 & 35.06 & \multicolumn{1}{|c}{\cellcolor[HTML]{FFFBCF}44.51} \\
             & Fine-Tuning & \multicolumn{1}{|c}{\underline{76.53}} & \textbf{85.24} & \underline{47.38} & \multicolumn{1}{|c}{42.78} & 42.92 & 52.48 & 15.54 & 55.38 & 38.37 & \multicolumn{1}{|c}{\cellcolor[HTML]{FFFBCF}\underline{50.73}} \\
             & LLM-AMT & \multicolumn{1}{|c}{52.91} & 66.63 & 45.08 & \multicolumn{1}{|c}{44.60} & \textbf{45.39} & \underline{53.07} & \underline{16.22} & \underline{60.20} & \underline{40.29} & \multicolumn{1}{|c}{\cellcolor[HTML]{FFFBCF}47.16} \\
             & \ours & \multicolumn{1}{|c}{\textbf{77.01}} & \underline{84.16} & \textbf{52.33} & \multicolumn{1}{|c}{\textbf{52.75}} & \underline{45.19} & \textbf{55.69} & \textbf{17.79} & \textbf{62.90} & \textbf{44.82} & \multicolumn{1}{|c}{\cellcolor[HTML]{FFFBCF}\textbf{54.74}} \\
             \midrule
             \multirow{5}{*}{Qwen2.5-14B} & Direct Response & \multicolumn{1}{|c}{50.01} & 65.23 & 42.85 & \multicolumn{1}{|c}{56.93} & 71.60 & 45.63 & 11.06 & 43.17 & 26.58 & \multicolumn{1}{|c}{\cellcolor[HTML]{FFFBCF}45.89} \\
             & Vanilla RAG & \multicolumn{1}{|c}{\underline{54.88}} & 75.60 & 42.06 & \multicolumn{1}{|c}{60.38} & 79.46 & 45.69 & 12.28 & 54.00 & 28.05 & \multicolumn{1}{|c}{\cellcolor[HTML]{FFFBCF}50.27} \\
             & Fine-Tuning & \multicolumn{1}{|c}{\textbf{55.87}} & \textbf{85.52} & \underline{47.64} & \multicolumn{1}{|c}{54.29} & \underline{80.41} & 46.67 & 12.30 & \underline{66.32} & \underline{43.66} & \multicolumn{1}{|c}{\cellcolor[HTML]{FFFBCF}\underline{54.75}} \\
             & LLM-AMT & \multicolumn{1}{|c}{51.48} & 77.11 & 43.09 & \multicolumn{1}{|c}{\underline{62.42}} & 78.78 & \underline{47.44} & \underline{13.73} & 60.03 & 38.81 & \multicolumn{1}{|c}{\cellcolor[HTML]{FFFBCF}52.54} \\
             & \ours & \multicolumn{1}{|c}{54.34} & \underline{80.19} & \textbf{48.36} & \multicolumn{1}{|c}{\textbf{68.32}} & \textbf{84.03} & \textbf{48.03} & \textbf{14.45} & \textbf{72.80} & \textbf{46.30}  & \multicolumn{1}{|c}{\cellcolor[HTML]{FFFBCF}\textbf{57.43}} \\
             \midrule
             \multirow{5}{*}{Qwen2.5-32B} & Direct Response & \multicolumn{1}{|c}{16.23} & 87.07 & 66.44 & \multicolumn{1}{|c}{\textbf{68.66}} & 80.19 & 43.59 & 15.06 & 59.67 & 31.33 & \multicolumn{1}{|c}{\cellcolor[HTML]{FFFBCF}52.03} \\
             & Vanilla RAG & \multicolumn{1}{|c}{19.33} & 89.30 & 67.63 & \multicolumn{1}{|c}{67.06} & \underline{83.85} & 46.63 & 17.49 & 66.10 & 40.59 & \multicolumn{1}{|c}{\cellcolor[HTML]{FFFBCF}55.33} \\
             & Fine-Tuning & \multicolumn{1}{|c}{\textbf{25.57}} & \underline{89.97} & 66.17 & \multicolumn{1}{|c}{66.69} & 82.40 & \textbf{52.58} & 15.71 & \underline{70.20} & 45.08 & \multicolumn{1}{|c}{\cellcolor[HTML]{FFFBCF}\underline{57.15}} \\
             & LLM-AMT & \multicolumn{1}{|c}{19.38} & 88.08 & \underline{68.33} & \multicolumn{1}{|c}{\underline{68.61}} & 82.06 & 46.82 & \underline{17.90} & 66.75 & \underline{45.40} & \multicolumn{1}{|c}{\cellcolor[HTML]{FFFBCF}55.93} \\
             & \ours & \multicolumn{1}{|c}{\underline{24.43}} & \textbf{90.01} & \textbf{69.09} & \multicolumn{1}{|c}{68.36} & \textbf{84.95} & \underline{50.40} & \textbf{18.54} & \textbf{72.51} & \textbf{55.36} & \multicolumn{1}{|c}{\cellcolor[HTML]{FFFBCF}\textbf{59.29}} \\
            \midrule
            \multirow{5}{*}{LLaMA3.1-70B} & Direct Response & \multicolumn{1}{|c}{46.43} & 58.36 & 62.33 & \multicolumn{1}{|c}{66.81} & 71.33 & 65.43 & 23.43 & 72.59 & 43.54 & \multicolumn{1}{|c}{\cellcolor[HTML]{FFFBCF}56.69} \\
             & Vanilla RAG & \multicolumn{1}{|c}{62.66} & 77.91 & 66.63 & \multicolumn{1}{|c}{68.78} & 76.95 & 66.40 & 25.95 & \underline{77.70} & 44.88 & \multicolumn{1}{|c}{\cellcolor[HTML]{FFFBCF}63.10} \\
             & Fine-Tuning & \multicolumn{1}{|c}{\textbf{87.17}} & \textbf{86.21} & \underline{71.65} & \multicolumn{1}{|c}{74.72} & 79.46 & 68.11 & 24.18 & 76.41 & \underline{47.23} & \multicolumn{1}{|c}{\cellcolor[HTML]{FFFBCF}\underline{68.35}} \\
             & LLM-AMT & \multicolumn{1}{|c}{79.18} & 68.59 & 70.12 & \multicolumn{1}{|c}{\textbf{79.05}} & \underline{80.74} & \underline{68.90} & \underline{26.53} & 77.00 & 45.87 & \multicolumn{1}{|c}{\cellcolor[HTML]{FFFBCF}66.22} \\
             & \ours & \multicolumn{1}{|c}{\underline{86.37}} & \underline{84.58} & \textbf{73.36} & \multicolumn{1}{|c}{\underline{78.24}} & \textbf{82.82} & \textbf{70.18} & \textbf{26.84} & \textbf{78.60} & \textbf{49.42} & \multicolumn{1}{|c}{\cellcolor[HTML]{FFFBCF}\textbf{70.04}} \\
            \bottomrule
        \end{tabular}
    }
\end{table*}

\subsection{Experimental Setup}
\label{subsec:experimental_setup}

\textbf{Model Details.} 
We employ the open-sourced LLMs from LLaMA~\cite{touvron2023llama2,dubey2024llama} and Qwen~\cite{yang2024qwen2} series as our target models for evaluation. 
We assessed models including 
LLaMA3.1-8B, 
Qwen2.5-14B, Qwen2.5-32B and LLaMA3.1-70B, scaling from 8B to 70B. 
The default setting of context window for our main experiments is 4K, with an in-depth scaling analysis presented in \Cref{sec:ablation}. 

\textbf{Datasets.} 
We have selected eight medical datasets including PubMedQA~\cite{jin2019pubmedqa}, MedQA-USMLE, MedQA-MCMLE~\cite{jin2020disease}, MedMCQA~\cite{pal2022medmcqa}, MMLU-Med~\cite{hendryckstest2021}, 
NEJMQA~\cite{katz2024gpt}, MedXpertQA~\cite{zuo2025medxpertqa}, and RareArena~\cite{THUMedInfo_RareArena}, 
covering both standard and real-world clinical scenarios. 
We use accuracy as evaluation metrics. 
More details about the datasets can be found in \Cref{appendix_subsec:medical_datasets}. 


\textbf{Implementation.} 
We constructed the medical knowledge corpus by establishing FAISS vector library~\cite{johnson2019billion}. 
Experiments related to model training were conducted based on full-parameter fine-tuning, during which we utilized a learning rate scheduler featuring linear warm-up and cosine decay, peaking at a learning rate of 2e\mbox{-}5, alongside a warmup ratio of 0.03, a weight decay of 0.0 and a batch size of 128 for 3 epochs.
We conducted all training and evaluation experiments on NVIDIA RTX H800 GPUs with 80G memory.

\subsection{Baselines}
\label{subsec:baselines}

We compare \ours with the following baselines:
(1) \textbf{\textit{Frontier Models}} contain 
GPT-4o~\cite{hurst2024gpt}, 
Claude3.5-Sonnet~\cite{claude_anthropic_2024} and DeepSeek-V3~\cite{liu2024deepseek}. 
(2) \textbf{\textit{Open-Sourced Medical Models}} include PMC-LLaMA-7B, PMC-LLaMA-13B~\cite{wu2024pmc}, MEDITRON-7B and MEDITRON-70B~\cite{chen2023meditron}. 
(3) The simplest baseline is \textbf{\textit{Direct Response}}, where the model answer medical questions directly without the aid of external knowledge bases or dataset fine-tuning. 
(4) \textbf{\textit{Vanilla RAG}}~\cite{lewis2020retrieval} utilizes raw queries for evidence searching, and the retrieved documents are then directly integrated into the generation process without any further manipulation. 
(5) \textbf{\textit{Fine-Tuning}} leverages medical datasets to further train the model under supervised conditions. 
Here we employ two strategies: \textit{within-dataset fine-tuning}, where the datasets for training and evaluation are derived from different parts of the same data corpus, and \textit{cross-dataset fine-tuning}, where the model is fine-tuned on one medical dataset (e.g., MedMCQA) and then evaluated on different datasets (e.g., PubMedQA). 
(6) \textbf{\textit{LLM-AMT}}~\cite{wang2024augmenting} is a dedicated process tailored for biomedical question answering, which includes typical modules such as query augmenter, hybrid retriever, knowledge refiner, etc.
Details are discussed in \Cref{appendix_subsec:baselines}.

\subsection{Main Results}
\label{subsec:results}

We have performed evaluations to validate the efficiency of our \ours on open-sourced models across different parameter scales. 
The main results of baselines and \ours are demonstrated in \Cref{tab:main_results}, and we summarize the observations below.


\textbf{\ours is effective across different models.} 
\Cref{tab:main_results} shows that the incorporation of external knowledge bases significantly enhances the model's ability to address medical queries, where even the most basic \textit{vanilla RAG} method depicts an average enhancement of 13.10\% over \textit{direct responses}. 
Furthermore, \ours provides an added layer of the enhancement by adhering to the EBM process, which outperforms all baselines across benchmarks, achieving an average improvement of 28.10\% over the \textit{direct response} strategy. 
Notably, for lightweight models such as LLaMA3.1-8B, \ours demonstrates increases of 61.43\%. We surmise that this is due to the fact that while lightweight models inherently lack comprehensive domain-specific medical knowledge, they possess the capability to efficiently read and identify information from external medical documents. Consequently, effective augmentation from external knowledge substantially bolsters the models' capacity to tackle medical-domain questions. 
Compared with medical-specific models of the same parameter scale, models equipped with Med-R$^2$ demonstrate superior performance in medical tasks. 
Moreover, LLaMA3.1-70B + \ours has the potential to surpass frontier models under medical scenarios, achieving average improvements over GPT-4o, Claude3.5-Sonnet and DeepSeek-V3 by 1.05\%, 6.14\% and 1.91\%. 

\textbf{\ours shows superiority compared to fine-tuning.} 
From \Cref{tab:main_results}, we find that \ours stands out as the only approach among those leveraging external knowledge that surpasses the average performance of \textit{fine-tuning} methods. Specifically, \ours exhibits nearly equivalent performance to fine-tuning strategies in \textit{within-dataset training}, yet it significantly outperforms in \textit{cross-dataset training}, achieving an enhancement of 7.61\% and an overall capability improvement of 4.55\%. One contributing factor is that during within-dataset fine-tuning, the training and testing datasets are of the same origin, thus a model trained on homogeneous data could achieve substantial performance gains on the test set. Conversely, in cross-dataset fine-tuning, the heterogeneity between the training and testing datasets more rigorously assesses the model's ability of generalization. Under these circumstances, the utilization of a comprehensive external medical knowledge base and the effective retrieval and extraction of pertinent information becomes particularly crucial. 
Additionally, considering that fine-tuning necessitates additional training time and computational resources, \ours emerges as a more efficient approach for enhancing the model's performance on medical domain-related issues.

\section{Ablations and Analysis Across Scales}
\label{sec:ablation}

We further analyze the impact of \textit{model scale} and \textit{context window length} on \ours, and provide rationale for incorporating the CoT sequences of models into the retrieval process starting from the second iteration. 
Additionally, we have also ablate the components of \ours in \Cref{tab:component_analysis} to identify the contribution of each module.

\begin{figure}[th]
    \begin{center}
    \centerline{\includegraphics[width=\columnwidth]{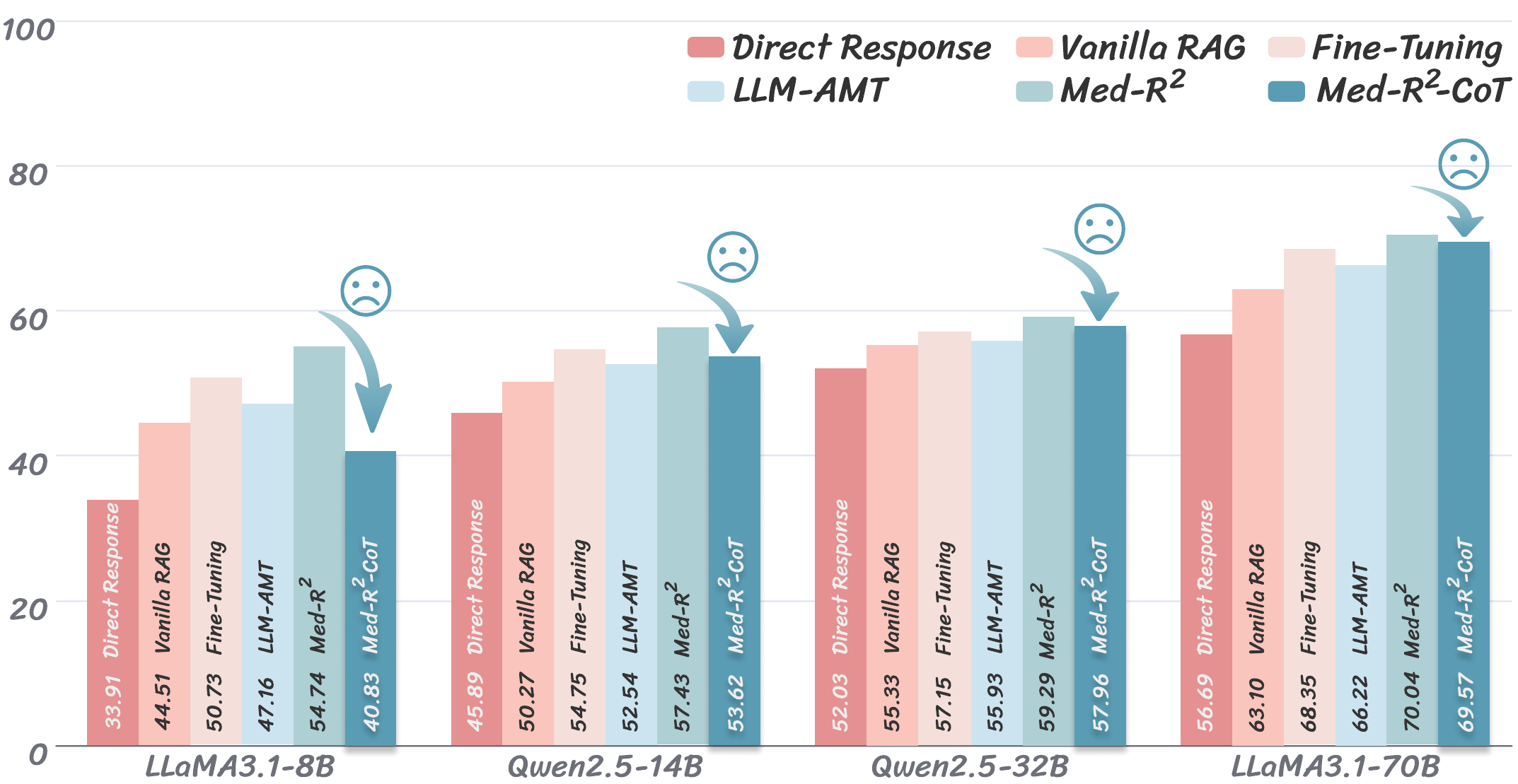}}
    \caption{Average evaluation results of baselines from \Cref{tab:main_results} and Med-R$^2$-CoT. Compared to Med-R$^2$, Med-R$^2$-CoT involves the immediate incorporation of models' CoT sequence into the query reformulator for retrieval during the initial round.}
    \label{fig:cot_barchart}
    \end{center}
    \vskip -0.2in
\end{figure}

\textbf{Integrating CoT at the onset may bring adverse effects.} 
We ablate the components of query reformulator $q_{EBM} \left[+q_{CoT}\right]$ through integrating the chain-of-thought (CoT) sequences generated by models into the initial evidence retrieval phase, which is denoted as \textbf{\textit{Med-R$^2$-CoT}}. 
\Cref{fig:cot_barchart} presents the medical task performance of Med-R$^2$-CoT and all baselines included in our main experiments. It is observed that Med-R$^2$-CoT exhibits a decline in performance compared to Med-R$^2$, with the disparity increasing as the model parameter scale decreases. Notably, at the 8 billion parameter level (e.g., LLaMA3.1-8B), the performance of Med-R$^2$-CoT is even inferior to that of the \textit{vanilla RAG} strategy. 
We hypothesize that one contributing factor is that the lightweight models' less solid grasp of medical knowledge. As a result, in the absence of external medical knowledge, models of these scales may struggle to align the direction of their thinking with the original query, and thus, the generated CoT sequences may even negatively impact the retrieval effectiveness. However, this phenomenon is somewhat mitigated as the model parameter scale increases.

\begin{figure}[th]
    \begin{center}
    \centerline{\includegraphics[width=\columnwidth]{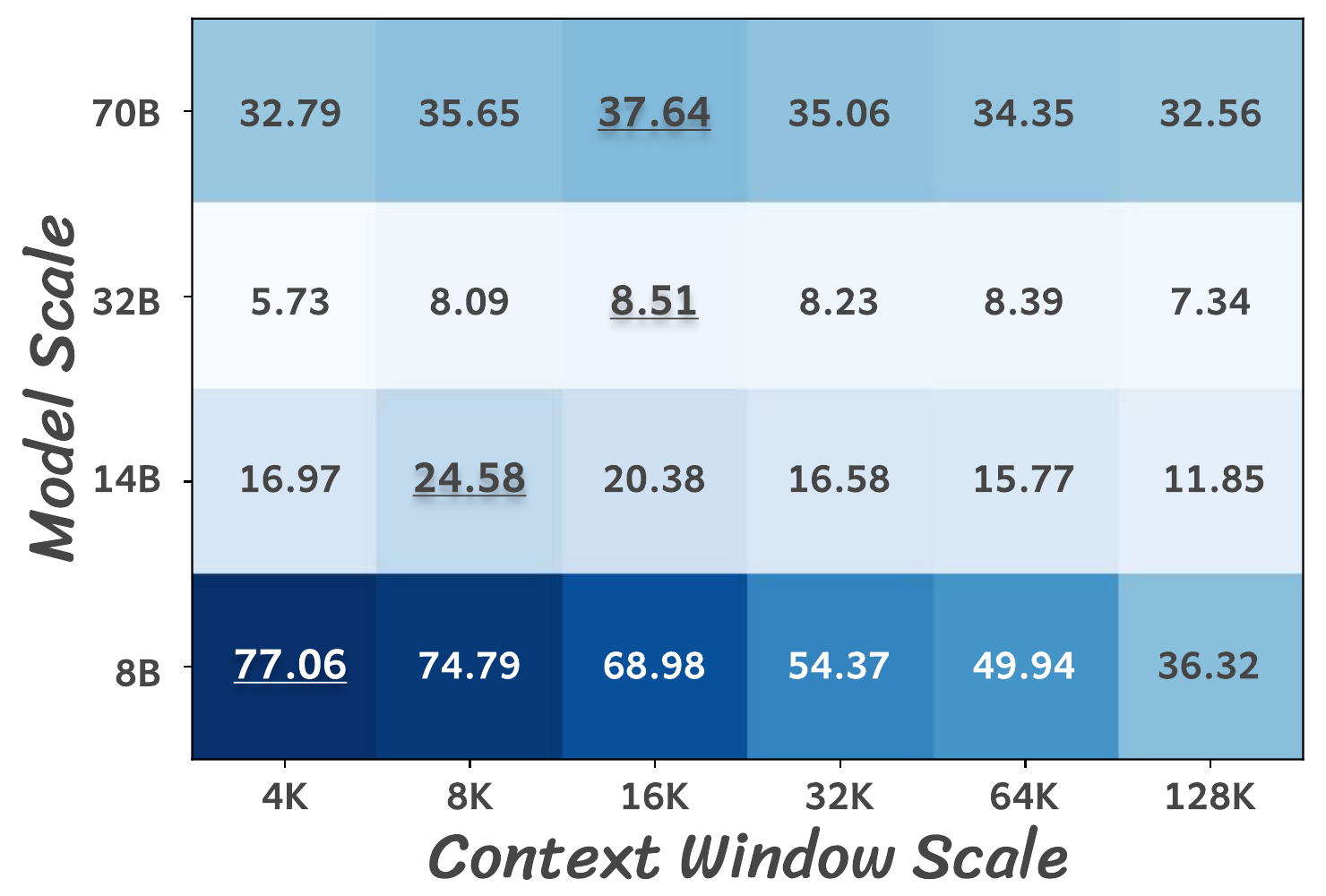}}
    \caption{The percentage increase of performance achieved by \ours over \textit{direct responses} across various context window sizes and model scales.}
    \label{fig:context_window_scaling}
    \end{center}
    \vskip -0.2in
\end{figure}

\begin{table*}[ht]
    \centering
     \caption{\label{tab:component_analysis}
    Module analysis of \ours. We sequentially integrate various modules onto the \textit{vanilla RAG} systems to conduct comparative analyses. 
    We highlight the optimal performance values for models with \colorbox[HTML]{F7EE90}{single} and \colorbox[HTML]{F0DE26}{dual} component additions. 
} 
    \resizebox{\textwidth}{!}{
        \begin{tabular}{clcccccc}
            \toprule
                Model & \multicolumn{1}{|c}{Method} & \multicolumn{1}{|c}{MedQA-USMLE} & MedQA-MCMLE & MedMCQA &  \multicolumn{1}{|c}{PubMedQA} &  \multicolumn{1}{c|}{MMLU-Med} & Average \\
            \midrule
            \multirow{9}{*}{LLaMA3.1-8B} & \multicolumn{1}{|c}{Direct Response} & \multicolumn{1}{|c}{31.16} & 41.45 & 30.02 & \multicolumn{1}{|c}{36.17} & 37.12 & \multicolumn{1}{|c}{\cellcolor[HTML]{FFFBCF}35.18} \\
            \cline{2-8}
            & \multicolumn{1}{|c}{Vanilla RAG} & \multicolumn{1}{|c}{55.80} & 59.38 & 35.91 & \multicolumn{1}{|c}{47.10} & 43.54& \multicolumn{1}{|c}{\cellcolor[HTML]{FFFBCF}48.35} \\
            \cline{2-8}
             & \multicolumn{1}{|c}{+ Query Reformulator} & \multicolumn{1}{|c}{62.47} & 72.85& 41.66& \multicolumn{1}{|c}{48.72} & 43.98& \multicolumn{1}{|c}{\cellcolor[HTML]{FFFBCF}53.94} \\
             & \multicolumn{1}{|c}{+ Evidence Reranker} & \multicolumn{1}{|c}{68.62} & 75.74& 43.84& \multicolumn{1}{|c}{49.97} & 43.76& \multicolumn{1}{|c}{\cellcolor[HTML]{F7EE90}56.34} \\
             & \multicolumn{1}{|c}{+ CoT Generator} & \multicolumn{1}{|c}{59.87} & 65.63& 39.45& \multicolumn{1}{|c}{47.71} & 43.61& \multicolumn{1}{|c}{\cellcolor[HTML]{FFFBCF}51.25} \\
             & \multicolumn{1}{|c}{+ Query Reformulator, Evidence Reranker} & \multicolumn{1}{|c}{74.41} & 81.68& 47.71& \multicolumn{1}{|c}{53.96} & 44.23& \multicolumn{1}{|c}{\cellcolor[HTML]{F0DE26}60.40} \\
             & \multicolumn{1}{|c}{+ Query Reformulator, CoT Generator} & \multicolumn{1}{|c}{70.86} & 77.43& 44.83& \multicolumn{1}{|c}{51.75} & 44.02& \multicolumn{1}{|c}{\cellcolor[HTML]{FFFBCF}57.78} \\
             & \multicolumn{1}{|c}{+ Evidence Reranker, CoT Generator} & \multicolumn{1}{|c}{72.69} & 79.84& 45.12& \multicolumn{1}{|c}{52.25} & 44.00 & \multicolumn{1}{|c}{\cellcolor[HTML]{FFFBCF}58.78} \\
             \cline{2-8}
             & \multicolumn{1}{|c}{\ours (ours)} & \multicolumn{1}{|c}{77.01} & 84.16 & 52.33 & \multicolumn{1}{|c}{52.75} & 45.19 & \multicolumn{1}{|c}{\cellcolor[HTML]{FFFBCF}62.29} \\
             \midrule
             \multirow{9}{*}{Qwen2.5-14B} & \multicolumn{1}{|c}{Direct Response} & \multicolumn{1}{|c}{50.01} & 65.23 & 42.85 & \multicolumn{1}{|c}{56.93} & 71.60 & \multicolumn{1}{|c}{\cellcolor[HTML]{FFFBCF}57.32} \\
             \cline{2-8}
             & \multicolumn{1}{|c}{Vanilla RAG} & \multicolumn{1}{|c}{54.88} & 75.60 & 42.06 & \multicolumn{1}{|c}{60.38} & 79.46 & \multicolumn{1}{|c}{\cellcolor[HTML]{FFFBCF}62.48} \\
             \cline{2-8}
             & \multicolumn{1}{|c}{+ Query Reformulator} & \multicolumn{1}{|c}{54.93} & 78.96 & 45.43 & \multicolumn{1}{|c}{64.85} & 81.73 & \multicolumn{1}{|c}{\cellcolor[HTML]{FFFBCF}65.18} \\
             & \multicolumn{1}{|c}{+ Evidence Reranker} & \multicolumn{1}{|c}{55.03} & 78.83 & 45.97 & \multicolumn{1}{|c}{65.08} & 81.56 & \multicolumn{1}{|c}{\cellcolor[HTML]{F7EE90}65.29} \\
             & \multicolumn{1}{|c}{+ CoT Generator} & \multicolumn{1}{|c}{54.62} & 76.72 & 43.86 & \multicolumn{1}{|c}{62.13} & 80.02 & \multicolumn{1}{|c}{\cellcolor[HTML]{FFFBCF}63.47} \\
             & \multicolumn{1}{|c}{+ Query Reformulator, Evidence Reranker} & \multicolumn{1}{|c}{55.38} & 82.04 & 48.29 & \multicolumn{1}{|c}{67.02} & 83.17 & \multicolumn{1}{|c}{\cellcolor[HTML]{F0DE26}67.18} \\
             & \multicolumn{1}{|c}{+ Query Reformulator, CoT Generator} & \multicolumn{1}{|c}{54.77} & 80.58 & 47.61 & \multicolumn{1}{|c}{66.98} & 82.21 & \multicolumn{1}{|c}{\cellcolor[HTML]{FFFBCF}66.43} \\
             & \multicolumn{1}{|c}{+ Evidence Reranker, CoT Generator} & \multicolumn{1}{|c}{54.81} & 81.79 & 47.87 & \multicolumn{1}{|c}{67.48} & 82.95 & \multicolumn{1}{|c}{\cellcolor[HTML]{FFFBCF}66.98} \\
             \cline{2-8}
             & \multicolumn{1}{|c}{\ours (ours)}  & \multicolumn{1}{|c}{54.34} & 80.19 & 48.36 & \multicolumn{1}{|c}{68.32} & 84.03 & \multicolumn{1}{|c}{\cellcolor[HTML]{FFFBCF}67.05} \\
             \midrule
             \multirow{9}{*}{Qwen2.5-32B} & \multicolumn{1}{|c}{Direct Response} & \multicolumn{1}{|c}{16.23} & 87.07 & 66.44 & \multicolumn{1}{|c}{68.66} & 80.19 & \multicolumn{1}{|c}{\cellcolor[HTML]{FFFBCF}63.72} \\
             \cline{2-8}
             & \multicolumn{1}{|c}{Vanilla RAG} & \multicolumn{1}{|c}{19.33} & 89.30 & 67.63 & \multicolumn{1}{|c}{67.06} & 83.85 & \multicolumn{1}{|c}{\cellcolor[HTML]{FFFBCF}65.43} \\
             \cline{2-8}
             & \multicolumn{1}{|c}{+ Query Reformulator} & \multicolumn{1}{|c}{20.14} & 89.42 & 68.72 & \multicolumn{1}{|c}{68.41} & 83.97 & \multicolumn{1}{|c}{\cellcolor[HTML]{FFFBCF}66.13} \\
             & \multicolumn{1}{|c}{+ Evidence Reranker} & \multicolumn{1}{|c}{20.83} & 89.88 & 68.88 & \multicolumn{1}{|c}{68.47} & 83.99 & \multicolumn{1}{|c}{\cellcolor[HTML]{F7EE90}66.41} \\
             & \multicolumn{1}{|c}{+ CoT Generator} & \multicolumn{1}{|c}{19.78} & 88.69 & 67.94 & \multicolumn{1}{|c}{67.72} & 83.92 & \multicolumn{1}{|c}{\cellcolor[HTML]{FFFBCF}65.61} \\
             & \multicolumn{1}{|c}{+ Query Reformulator, Evidence Reranker} & \multicolumn{1}{|c}{23.04} & 89.25 & 70.35 & \multicolumn{1}{|c}{68.89} & 84.46 & \multicolumn{1}{|c}{\cellcolor[HTML]{F0DE26}67.20} \\
             & \multicolumn{1}{|c}{+ Query Reformulator, CoT Generator} & \multicolumn{1}{|c}{21.97} & 89.56 & 69.06 & \multicolumn{1}{|c}{68.56} & 84.08 & \multicolumn{1}{|c}{\cellcolor[HTML]{FFFBCF}66.65} \\
             & \multicolumn{1}{|c}{+ Evidence Reranker, CoT Generator} & \multicolumn{1}{|c}{23.01} & 89.75 & 69.75 & \multicolumn{1}{|c}{69.00} & 84.39 & \multicolumn{1}{|c}{\cellcolor[HTML]{FFFBCF}67.18} \\
             \cline{2-8}
             & \multicolumn{1}{|c}{\ours (ours)} & \multicolumn{1}{|c}{24.43} & 90.01 & 69.09 & \multicolumn{1}{|c}{68.36} & 84.95 & \multicolumn{1}{|c}{\cellcolor[HTML]{FFFBCF}67.37} \\
             \midrule
             \multirow{9}{*}{LLaMA3.1-70B} & \multicolumn{1}{|c}{Direct Response} & \multicolumn{1}{|c}{46.43} & 58.36 & 62.33 & \multicolumn{1}{|c}{66.81} & 71.33 & \multicolumn{1}{|c}{\cellcolor[HTML]{FFFBCF}61.05} \\
             \cline{2-8}
             & \multicolumn{1}{|c}{Vanilla RAG} & \multicolumn{1}{|c}{62.66} & 77.91 & 66.63 & \multicolumn{1}{|c}{68.78} & 76.95 & \multicolumn{1}{|c}{\cellcolor[HTML]{FFFBCF}70.59} \\
             \cline{2-8}
             & \multicolumn{1}{|c}{+ Query Reformulator} & \multicolumn{1}{|c}{70.82} & 80.64 & 69.72 & \multicolumn{1}{|c}{71.82} & 73.56 & \multicolumn{1}{|c}{\cellcolor[HTML]{FFFBCF}73.31} \\
             & \multicolumn{1}{|c}{+ Evidence Reranker} & \multicolumn{1}{|c}{73.96} & 80.78 & 70.87 & \multicolumn{1}{|c}{74.64} & 75.28 & \multicolumn{1}{|c}{\cellcolor[HTML]{F7EE90}75.11} \\
             & \multicolumn{1}{|c}{+ CoT Generator} & \multicolumn{1}{|c}{68.08} & 78.65 & 68.85 & \multicolumn{1}{|c}{70.09} & 73.72 & \multicolumn{1}{|c}{\cellcolor[HTML]{FFFBCF}71.88} \\
             & \multicolumn{1}{|c}{+ Query Reformulator, Evidence Reranker} & \multicolumn{1}{|c}{80.41} & 84.09 & 73.01 & \multicolumn{1}{|c}{77.80} & 80.68 & \multicolumn{1}{|c}{\cellcolor[HTML]{F0DE26}79.20} \\
             & \multicolumn{1}{|c}{+ Query Reformulator, CoT Generator} & \multicolumn{1}{|c}{78.62} & 82.56 & 72.71 & \multicolumn{1}{|c}{76.06} & 78.56 & \multicolumn{1}{|c}{\cellcolor[HTML]{FFFBCF}77.70} \\
             & \multicolumn{1}{|c}{+ Evidence Reranker, CoT Generator} & \multicolumn{1}{|c}{81.93} & 82.97 & 72.88 & \multicolumn{1}{|c}{76.52} & 81.62 & \multicolumn{1}{|c}{\cellcolor[HTML]{FFFBCF}79.18} \\
             \cline{2-8}
             & \multicolumn{1}{|c}{\ours (ours)} & \multicolumn{1}{|c}{86.37} & 84.58 & 73.36 & \multicolumn{1}{|c}{78.24} & 82.82 & \multicolumn{1}{|c}{\cellcolor[HTML]{FFFBCF}81.07} \\
            \bottomrule
        \end{tabular}
    }
\end{table*}

\textbf{Effect of context window scale on model's performance.} 
We compared the performance of \ours across models with varying parameter scales and different context window lengths on medical tasks, 
and then plotted heatmaps illustrating the percentage improvement of \ours over \textit{direct responses}, as depicted in \Cref{fig:context_window_scaling}. 
It reveals that \ours exhibits an optimal context window length for enhancing the model's medical performance, which increases with the growth of model parameter size. 
Concurrently, the enhancement of \ours follows a trend of initial decline followed by an increase with the escalation of model scale. 
Specifically:

\begin{itemize}[leftmargin=*]
    \item For models of 8B parameters, the most pronounced enhancement was observed at a 4K context window, but this benefit diminished sharply as the context length increased.
    \item For 14B models, the 8K length stands out, where a measurable decrease in performances is observed as the context window expanded. Moreover, the improvement provided by \ours at this scale is \textit{the most modest} compared to others.
    \item Models with 32B and 70B parameters achieved optimal performance at a 16K context length, demonstrating relatively stable improvement across various context window lengths.
\end{itemize}


We hypothesize that as the context length increases, the role of Med-R$^2$'s reranker diminishes since most retrieved documents are fed into the same context window of the model. At this point, lightweight models, particularly those around 8B, which may not have a solid grasp of medical knowledge, and overly long sequences could reduce the model's efficiency in extracting key information, potentially leading to the generation of hallucinations and adversely affecting the model's medical performance. However, the addition of other modules, such as the query reformulator, can help improve the precision of knowledge retrieval, thereby mitigating this negative impact to some extent. In contrast, models with 32B and higher scales exhibit greater robustness, and the impact of increasing the window length on Med-R$^2$'s effectiveness is less pronounced.




\textbf{Module Impact Analysis.} 
We have analyzed the contributions of various modules in \ours to the model's performance in the medical domain using a default context window size of 4K, 
sequentially incorporating components into the vanilla RAG framework. 
As shown in \Cref{tab:component_analysis}, we decompose the modules into three components: \textbf{query reformulator}, \textbf{evidence reranker}, and \textbf{CoT generator}. 
Overall, the evidence reranker contributed the most to models' performances among the individual components. When combined with the query reformulator, the performance gains were even more pronounced, showing a synergistic effect. The addition of the CoT generator further enhanced the model’s ability to effectively utilize retrieved medical documents, providing substantial added value. 

\section{Discussion}
\label{sec:conclusion}


In this study, we follow the Evidence-Based Medicine (EBM) process to design a novel LLM physician framework, which effectively leverages the retrieval, filtering, and reasoning processes inherent to EBM. 
Experiments demonstrate that \ours holds superior advantages over existing strategies, while also reducing the substantial computational costs associated with model training. 
Furthermore, our analysis of model scale and context window size also highlights the scaling capabilities of Med-R$^2$. 
By scaling the model parameters and adjusting the context window length, we demonstrate the robustness of \ours in achieving optimal performance across a broad spectrum of medical tasks. 

\clearpage

\bibliographystyle{ACM-Reference-Format}
\bibliography{www2026.bib}


\appendix


\renewcommand{\contentsname}{Appendix}
\tableofcontents
\addtocontents{toc}{\protect\setcounter{tocdepth}{2}}
\setcounter{tocdepth}{3}


\section{Evidence-Based Medicine (EBM)}
\label{sec:EBM}

Evidence-Based Medicine (EBM) is defined as the conscientious, explicit, and judicious application of the best current evidence in making decisions regarding the care of individual patients. This practice entails the integration of individual clinical expertise with the most reliable external clinical evidence derived from systematic research~\cite{sackett1996evidence}. 
The EBM process typically includes five stages, \textit{question formulating}, \textit{evidence searching}, \textit{evidence appraising}, \textit{evidence applying} and \textit{effect assessing}, which aims to make the best possible health care decision through iterative improvements. 
To utilize ``best evidence'', researchers assess the quality of trials by determining the grading system based on the likelihood that the methods used and the results obtained are less prone to bias and more reliable. 
The \textbf{\textit{hierarchy of evidence}} guides the clinical decision-making, since not all evidence is created equal, as described in \Cref{fig:hierarchy_of_evidence}. 
The evidence hierarchy establishes the priority of references, particularly when conflicting facts are present within the retrieved evidence.
The construction and rationale behind the hierarchical structure of evidence grading is outlined from the highest to the lowest levels:

\begin{itemize}[leftmargin=*] 
    \item \textbf{Systematic Reviews/Meta-Analyses (SR/MA)} represent \textit{the highest} tier of evidence. These assessments evaluate the consistency and risk of bias across all research findings within the medical domain, demonstrating the overall effect of exposures. 
    \item \textbf{Randomized Controlled Trials (RCTs)} constitute \textit{the second-highest} level of evidence. These trials aim to minimize confounding biases and examine the causal relationships between intervention measures and outcomes across groups. 
    \item \textbf{Cohort Studies} fall into \textit{the third-highest} category of evidence. Both retrospective and prospective cohort studies are prone to various biases. 
    Prospective cohort studies are considered more reliable, less susceptible to information biases (selection, misclassification, recall), and can establish temporal associations (outcomes following exposure). However, these cohort studies may suffer from confounding biases, which is a major concern that can undermine the validity of their findings. 
    \item \textbf{Case-Control Studies} are a form of observational research and rank as \textit{the fourth-highest} level of evidence. These studies attempt to identify associations between outcomes and exposure to risk factors after the outcomes have occurred. Case-control studies are susceptible to selection, information, and confounding biases, reducing their credibility compared to cohort studies.
    \item \textbf{Individual Case Reports} are of \textit{the second-lowest} evidence level, essentially uncontrolled cohort studies lacking a comparison group. The absence of a control group affects the correlation between study variables, interventions factors and outcomes. 
    \item \textbf{Expert Opinion} is considered \textit{the lowest} level of evidence due to its high susceptibility to bias. Compared to other levels, experts tend to choose evidence that confirms their preconceived hypotheses, 
    potentially leading to conflicts of interest and a focus on a specific domain while overlooking broader contexts, 
    thereby introducing bias into their perspectives.
\end{itemize}

\section{Method Details}
\label{sec:appendix_method_details}

\subsection{Question Formulation Details}
\label{sec:appendix_question_formulation}

\subsubsection{Details of Query Classification}
\label{appendix_subsec:query_classification}

We categorize the medical queries along two orthogonal dimensions: \textit{Evidence-Based Medicine (EBM) categories} and \textit{general natural language question types}.

\textbf{Evidence-Based Medicine (EBM) Categories} \quad 
Nonprofessional queries may fail to clearly articulate the current medical symptoms, thereby hindering the effective retrieval of the necessary evidence. 
We classify medical queries into several types according to the categories of Evidence-Based Medicine (EBM) questions, including \textit{diagnosis}, \textit{therapy}, \textit{prognosis}, \textit{etiology}, \textit{prevention}, \textit{cost}, etc., and then conduct professional medical query reformulations based on their EBM categorization to emphasize specialized retrieval. Instructions for targeted augmentation of clinical queries are outlined in \Cref{tab:EBM_category}, where we employ Qwen2.5-72B-Instruct\footnote{https://huggingface.co/Qwen/Qwen2.5-72B-Instruct} fine-tuned on our 100 manually annotated samples as our question reformulator.

\textbf{General Natural Language Question Types} \quad 
It serves as a crucial reference for filtering and reranking the documents retrieved in response to a query, since the emphasis of the expected answer varies with different question types. We categorize queries into 12 natural language classes, as illustrated in \Cref{tab:query_category}, establishing a mapping between the types of questions and that of evidence to form a question-answer typology (depicted in \Cref{fig:query_doc_projection}).

\begin{figure*}[ht] 
    \centering
    \begin{subfigure}[b]{\columnwidth}
        \includegraphics[width=\columnwidth]{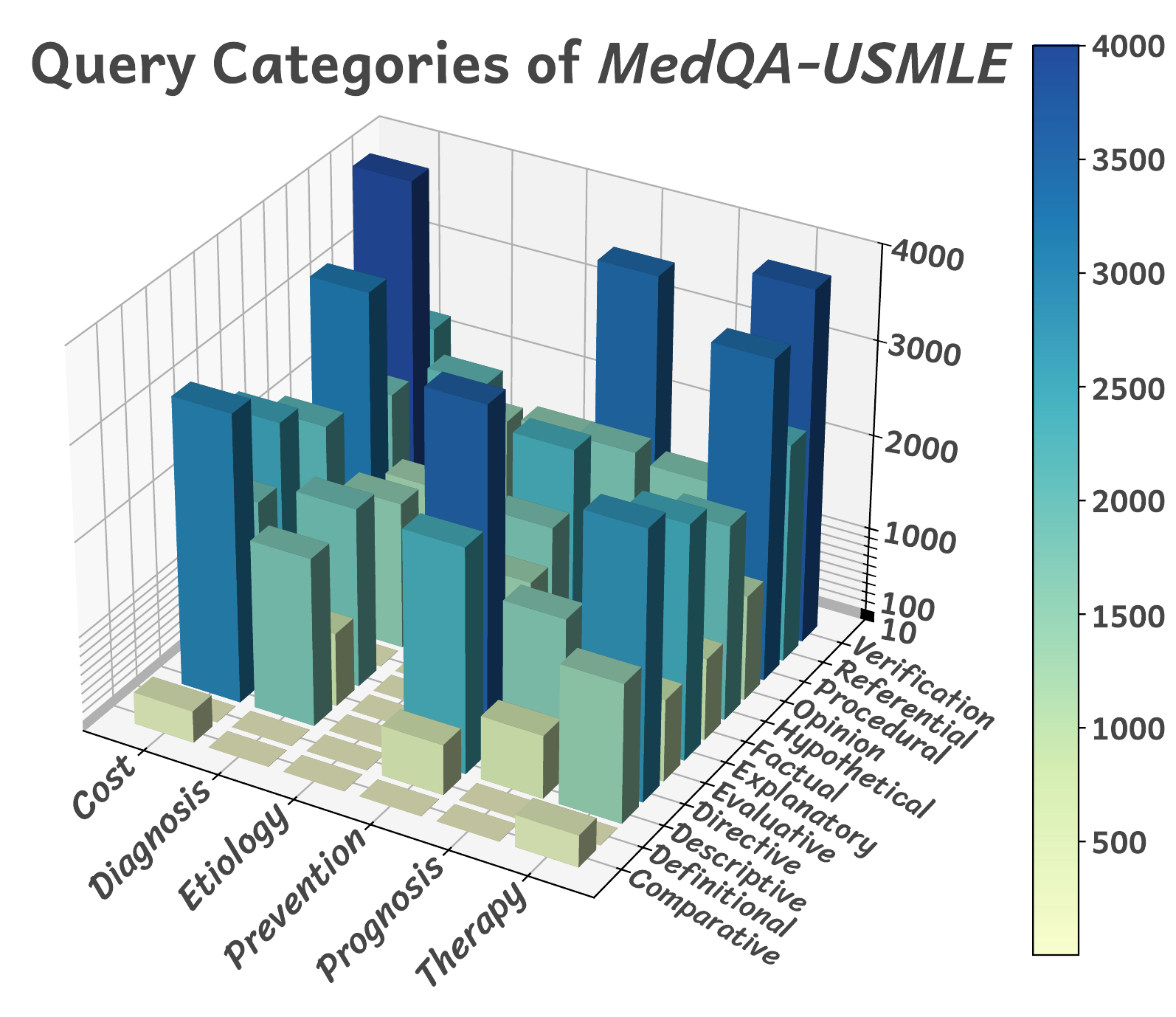}
        \caption{Query category of \textbf{MedQA-USMLE}.}
        \label{fig:MedQA-USMLE_query_category}
    \end{subfigure}
    \hfill
    \begin{subfigure}[b]{\columnwidth}
        \includegraphics[width=\columnwidth]{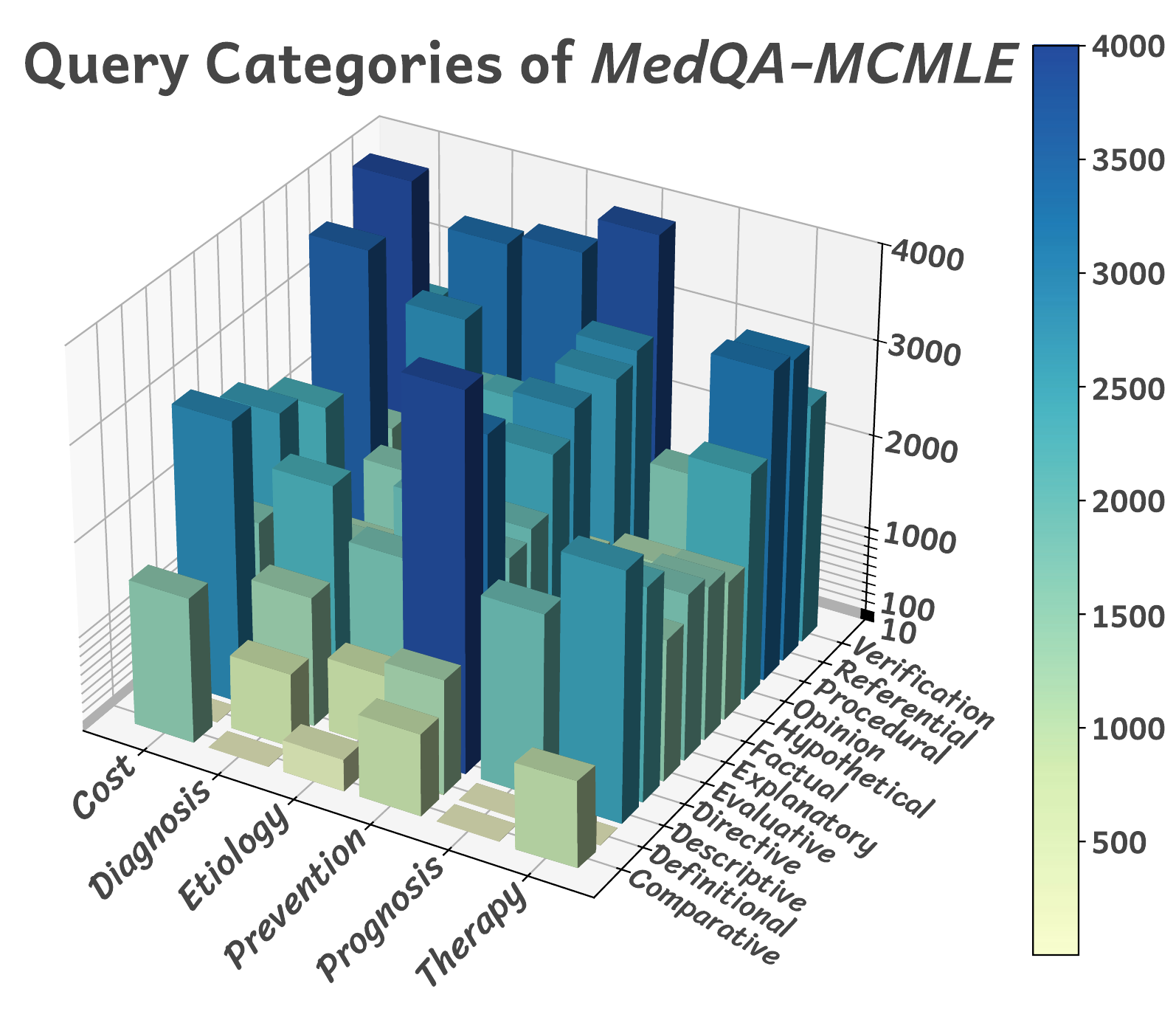}
        \caption{Query category of \textbf{MedQA-MCMLE}.}
        \label{fig:MedQA-MCMLE_query_category}
    \end{subfigure}


    \begin{subfigure}[b]{\columnwidth}
        \includegraphics[width=\columnwidth]{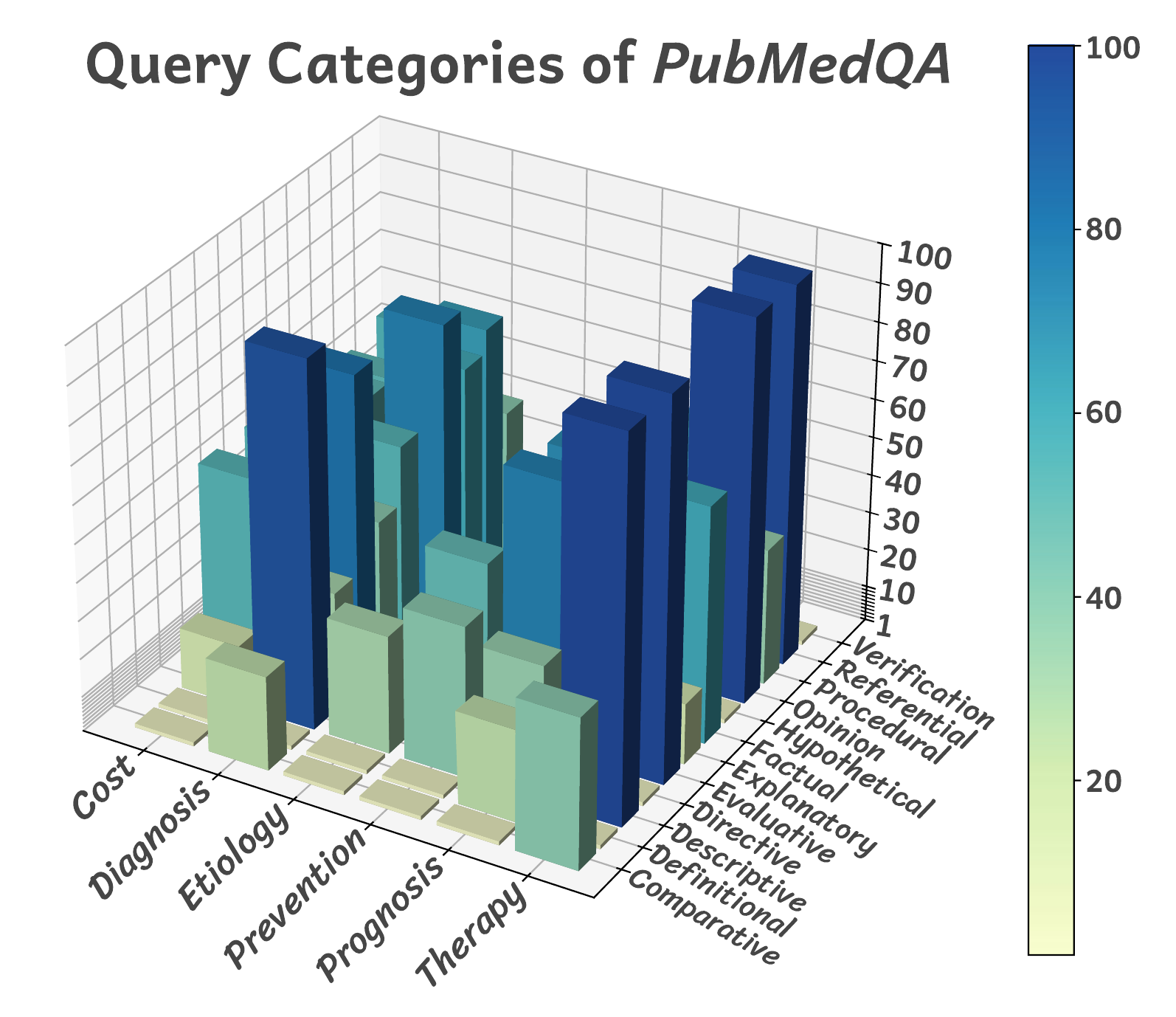}
        \caption{Query category of \textbf{PubMedQA}.}
        \label{fig:pubMedQA_query_category}
    \end{subfigure}
    \hfill
    \begin{subfigure}[b]{\columnwidth}
        \includegraphics[width=\columnwidth]{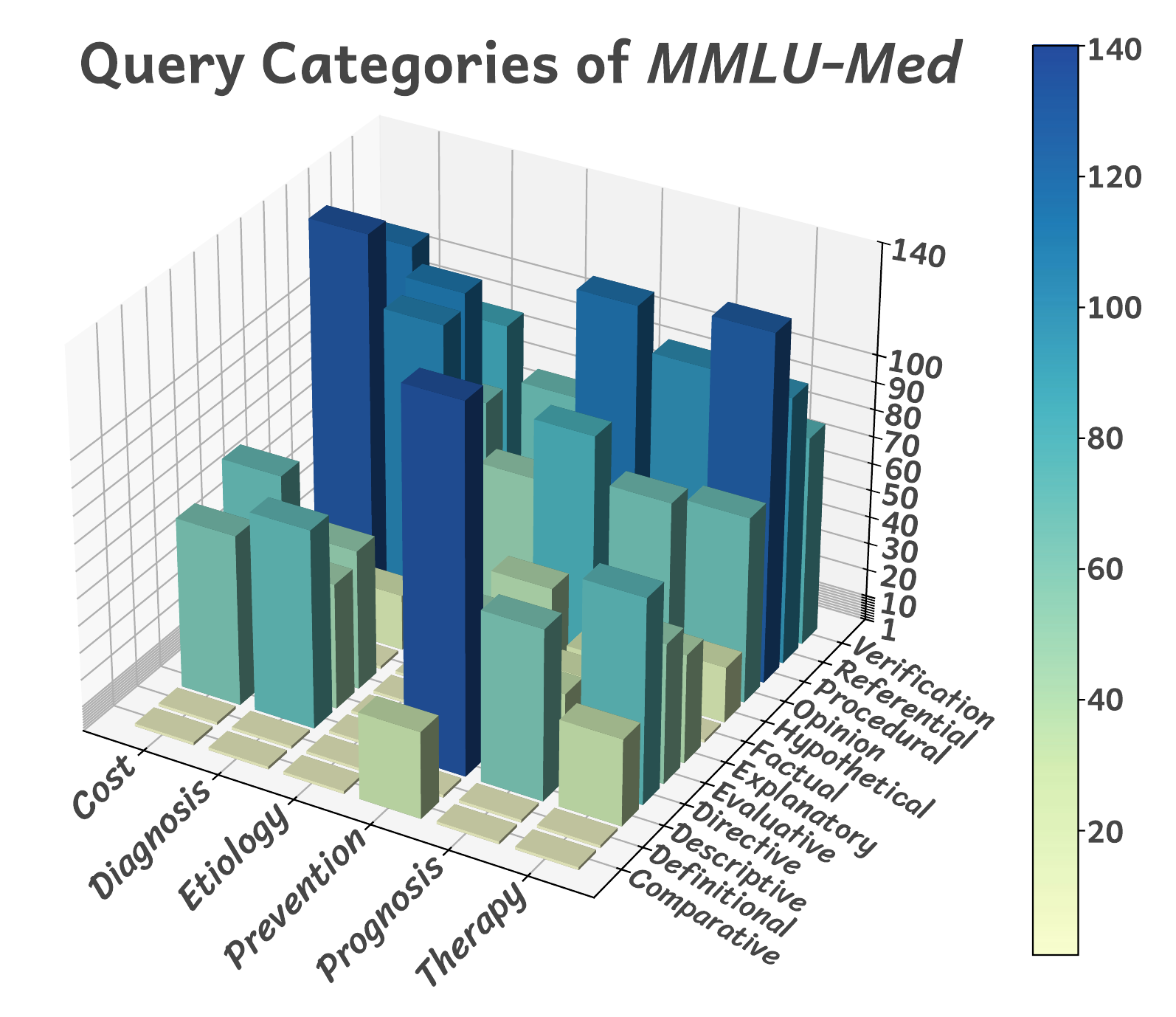}
        \caption{Query category of \textbf{MMLU-Med}.}
        \label{fig:MMLU_Med_query_category}
    \end{subfigure}

    \caption{Query categories of different medical datasets. We employ a 3D bar chart to represent the number of instances in each category within the two distinct classification systems of the current dataset. A logarithmic scale (base 10) on the z-axis, ranging from 1 to $z \, (z = 100, 140, 4000)$, is utilized to represent the wide range of values.}
    \label{fig:query_category_barchart}
\end{figure*}

\subsubsection{Details of Query Reformulator} 
\label{appendix_subsec:query_rewrite}

Following the instructions outlined in \Cref{tab:EBM_category}, we have manually annotated 120 samples (20 samples for each EBM category) to fine-tune the Qwen2.5-72B-Instruct model, and employed the trained model to professionally reformulate the queries within the medical datasets to enhance the subsequent retrieval of relevant evidence pertaining to the current query from the knowledge corpus. 




\subsection{Knowledge Corpus Details}
\label{sec:appendix_knowledge_corpus}


We organized the medical knowledge corpus from four data sources: \textit{academic papers}, \textit{entries}, \textit{books}, and \textit{guidelines}, as listed below:

\begin{table*}[t]
    \centering
    \caption{\label{tab:EBM_category} 
    Prompts for query reformulation of each category within the Evidence-Based Medicine (EBM) categories.}
    \resizebox{\textwidth}{!}{
        \begin{tabular}{c|c}
            \toprule
            Category & Instructions for Query Reformulation \\ 
            \midrule
            \textbf{\textit{Diagnosis}} & Specify the condition you need to diagnose and ask about the accuracy, sensitivity, or specificity of specific diagnostic tests. \\
            \midrule
            \textbf{\textit{Therapy}} & Specify the disease or symptom along with the therapy being considered, and inquire about its effectiveness, safety, or comparison with other therapies. \\
            \midrule
            \textbf{\textit{Prognosis}} & Specify the disease or condition and ask about long-term outcomes such as survival rates, recovery chances, or disease progression. \\
            \midrule
            \textbf{\textit{Etiology}} & Describe the health issue and ask about potential causes, including risk factors, pathogens, or genetic background. \\
            \midrule
            \textbf{\textit{Prevention}} & Specify the disease or health issue and ask about the effectiveness of preventive measures or recommendations. \\
            \midrule
            \textbf{\textit{Cost}} & Specify the medical intervention or service and ask about cost-effectiveness analyses, including direct and indirect costs and cost-effectiveness ratios. \\
            \bottomrule
        \end{tabular}
    }
    \vskip -0.2in
\end{table*}

\begin{table}[t]
    \centering
    \caption{\label{tab:query_category} 
    Explanations and descriptions for each category within the general question classification.}
    \resizebox{\columnwidth}{!}{
        \begin{tabular}{c|c}
            \toprule
            Category & Details of Description \\ 
            \midrule
            \textbf{\textit{Factual}} & Inquiring into specific and objective facts or data. \\
            \midrule
            \textbf{\textit{Referential}} & Seeking answers by referencing specific documents, resources, or other information. \\
            \midrule
            \textbf{\textit{Definition}} & Inquiring about the definition or explanation of a concept or entity. \\
            \midrule
            \textbf{\textit{Explanatory}} & Seeking explanations for the causes of phenomena, processes, or events. \\
            \midrule
            \textbf{\textit{Descriptive}} & Requesting a description of the characteristics, properties, and features of an entity. \\
            \midrule
            \textbf{\textit{Directive}} & Seeking guidance or recommendations. \\
            \midrule
            \textbf{\textit{Opinion}} & Pertaining to individual feelings, attitudes, or preferences. \\
            \midrule
            \textbf{\textit{Procedural}} & Inquiring about the specific steps to complete a particular task or activity. \\
            \midrule
            \textbf{\textit{Comparative}} & Inquiring comparison of the differences between two or more entities. \\
            \midrule
            \textbf{\textit{Evaluative}} & Assessing the validity or quality of a statement or viewpoint. \\
            \midrule
            \textbf{\textit{Verification}} & Confirming or verifying the authenticity or accuracy of certain information. \\
            \midrule
            \textbf{\textit{Hypothetical}} & Presenting a hypothetical scenario and requesting predictions of outcomes. \\
            \bottomrule
        \end{tabular}
    }
\end{table}

\begin{itemize}[leftmargin=*]
    \item \textbf{Academic Papers} \quad Academic literature provide valuable insights derived from the latest scientific investigations, offering a robust theoretical foundation for guiding clinical practice and informing public health decisions. We obtained the publicly available literature from PubMed Central (PMC)\footnote{https://pmc.ncbi.nlm.nih.gov/}, and exracted valid contents following the processing pipeline of ~\citet{hakala-etal-2016-syntactic}.
    \item \textbf{Entries} \quad Medical entries provide valuable information across multiple dimensions of healthcare, from direct patient care to cutting-edge research. We extracted and curated medical-related data from the Wikipedia dataset\footnote{https://huggingface.co/datasets/wikimedia/wikipedia} to build the entries. In the end, we got approximately 470k documents.
    \item \textbf{Books} \quad Medical textbooks are vital resources for medical knowledge retrieval, which can be consulted when faced with complex cases or when seeking updated knowledge about specific conditions. We gathered books from NCBI Bookshelf\footnote{https://www.ncbi.nlm.nih.gov/books/}, and collected 10k documents following the process of ~\citet{hakala-etal-2016-syntactic}.
    \item \textbf{Guidelines} \quad Clinical practice guidelines serve as essential tools in Evidence-Based Medicine (EBM), aiding healthcare providers in making informed decisions for diagnosis and treatment. We utilized the guideline data when training MEDITRON series from ~\citet{chen2023meditron}, and gained approximately 10k documents.
\end{itemize}

\subsection{Evidence Reranking Details}
\label{sec:appendix_evidence_reranking}

\subsubsection{Components of Fine-Grained Reranker}
\label{subsubsec:appendix_fine_grained_reranker_component}

We detail the scoring clarification and implementation for the fine-grained reranking phase of retrieved documents from \Cref{eq:fine_grained_reranking}.

\begin{itemize}[leftmargin=*]
    \item \textbf{Hierarchy of Evidence $f_h(x)$}: It should be clarified that for the \textit{guideline-sourced} data retrieved from the knowledge corpus, we assign the $f_h$ score directly to 7 ($e = 3$) based on the evidence level depicted in \Cref{fig:hierarchy_of_evidence}. It is because the grading system of evidence hierarchy explicitly designates a specific level for guideline data.
    \item \textbf{Usefulness $f_u(x)$}: \underline{\textit{\textbf{1)}}} Regarding the \textit{lightweight proxy models} we employed for scoring usefulness $f_u$ of the retrieved documents, for the LLaMA series, we used the Sheared-LLaMA-1.3B~\cite{xia2023sheared} as a lightweight reference model; for the Qwen series, we utilized Qwen-2.5-1.5B~\cite{yang2024qwen2}. \underline{\textit{\textbf{2)}}} Moreover, to \textit{avoid potential label leakage} when computing $f_u$ in \Cref{eq:f_u}, we have implemented strategies to ensure that the entire framework does not have access to ground truth answers (labels) during the deployment phase, where we employed frontier models' responses to serve as the ``reference answers'' for consultation throughout the framework's operation. 
    Specifically, we utilized the training subsets from MedQA-USMLE, MedQA-MCMLE~\cite{jin2020disease}, and MedMCQA~\cite{pal2022medmcqa} to fine-tune the Qwen2.5-72B-Instruct\footnote{https://huggingface.co/Qwen/Qwen2.5-72B-Instruct} model, resulting in model $M_F$. We replaced the labels with responses generated by model $M_F$ for the current medical question. 
    \item \textbf{General Document Category $f_g(x)$}: The calculation is based on the categorization and mapping to general document categories, as depicted in \Cref{alg:document_category_rerank}. For this categorization, advanced language models provide \textit{probabilities}. In addition, the relationship between the general question category and the general document category is \textit{multifaceted} (illustrated in ~\Cref{fig:query_doc_projection}), which further enhances the robustness of category assignment.
\end{itemize}

\begin{table}[ht]
    \centering
    \caption{\label{tab:document_category} 
    Explanations and descriptions for each category within the general document classification.}
    \resizebox{\columnwidth}{!}{
        \begin{tabular}{c|c}
            \toprule
            Category & Details of Description \\ 
            \midrule
            \textbf{\textit{Argumentation}} & Presenting a viewpoint or argument, potentially accompanied by supporting evidence. \\
            \midrule
            \textbf{\textit{Definition}} & Providing a clear definition of a term or concept. \\
            \midrule
            \textbf{\textit{Description}} & Describing the characteristics or attributes of an object or event. \\
            \midrule
            \textbf{\textit{Explanation}} & Explaining a concept, process, or cause. \\
            \midrule
            \textbf{\textit{Purpose}} & Elucidating the purpose or intent behind a particular action or event. \\
            \midrule
            \textbf{\textit{Narration}} & Providing a narrative account of an event, experience, or story. \\
            \midrule
            \textbf{\textit{Process}} & Describing a process or a sequence of steps. \\
            \midrule
            \textbf{\textit{Instruction}} & Providing steps or guidance for executing a task or operation. \\
            \midrule
            \textbf{\textit{Command}} & Conveying a request that requires the listener to take action. \\
            \midrule
            \textbf{\textit{Problem-Solving}} & Proposing methods or strategies for addressing specific issues. \\
            \midrule
            \textbf{\textit{Comparison}} & Comparing the similarities or differences between two or more entities. \\
            \midrule
            \textbf{\textit{Evaluation}} & Articulating a judgment on a particular subject or behavior. \\
            \midrule
            \textbf{\textit{Classification}} & Categorizing objects or concepts into specific categories systems. \\
            \midrule
            \textbf{\textit{Condition}} & Describing the assumptions under which a particular event occurs. \\
            \midrule
            \textbf{\textit{Prediction}} & Forecasting future events or trends. \\
            \midrule
            \textbf{\textit{Cause and Effect}} & Describing the causal relationships between events. \\
            \bottomrule
        \end{tabular}
    }
\end{table}

\subsubsection{Explanation for Equation Derivation}
\label{subsubsec:appendix_equation_derivation_explanation}

We have chosen \Cref{eq:fine_grained_reranking} in the form of \textit{multiplication} instead of additive scoring primarily for the following two reasons: 

\begin{itemize}[leftmargin=*]
    \item \textbf{Multi-Variable Synergy}: The multiplicative structure emphasizes the synergistic effect, preventing any single variable from dominating. It is necessary for all variables to reach a certain value \textit{simultaneously} to achieve the expected outcomes. For instance, if $f_h(x)$ or $f_g(x)$ approaches a lower bound (e.g., $f_g(x) \to 0$), even if the usefulness score $f_u(x)$ is extremely high, the final outcome $\mathcal{F}(x)$ will still tend towards 0.
    \item \textbf{Metric Self-Adaptation}: The value ranges of the variables differ significantly (e.g., $f_h(x) \in [1, 9]$, $f_g(x) \in (0, 1)$, $f_u(x) \in [0, +\infty)$). \Cref{eq:fine_grained_reranking} built upon the multiplicative formula naturally balances the scales of these variables. In contrast, the weighted summation approach requires manual normalization of variables.
\end{itemize}

\subsection{Evidence Assessment Details}
\label{sec:evidence_assessment}

Here, we provide the details of the iterative multi-document retrieval process, as illustrated in \Cref{alg:evidence_assessment}. 
\begin{itemize}[leftmargin=*]
    \item \textbf{Integration of CoT Sequences} \quad It is important to note that during the initial retrieval phase, we \textit{did not} incorporate the generated CoT sequences as part of the query reformulation $q_{EBM} \left[+q_{CoT}\right]$. Instead, this component was integrated in the \underline{\textit{second}} iteration, where we have shown the rationale in \Cref{sec:ablation}.
\end{itemize}

\section{Experiments Details}
\label{sec:appendix_experiments}
\subsection{Details of Medical Datasets}
\label{appendix_subsec:medical_datasets}

For MedQA-USMLE and MedQA-MCMLE, the original data is divided into three parts: train, dev, and test. We utilize the training part directly for model fine-tuning. The dev and test subsets are merged for evaluation, from which we selected 10 (or 11) instances as Chain-of-Thought (CoT) demonstration examples. 
Since the test portion of MedMCQA does not provide ground truth answers, we use the development set for evaluation. 
We employ MedQA-USMLE, MedQA-MCMLE, and MedMCQA datasets for \textit{within-dataset fine-tuning}. For instance, we train with the training subset of MedQA-USMLE and subsequently evaluate with its corresponding test partitions. 
We utilize the remaining datasets for \textit{cross-dataset fine-tuning} setting to test the model's generalizability on disparate data sources. 

\begin{table}[ht]
    \centering
    \caption{\label{tab:datasets_stats} 
    Overall statistics of datasets. We utilize MedQA-USMLE, MedQA-MCMLE and MedMCQA for \textit{within-dataset fine-tuning}, while others for \textit{cross-dataset fine-tuning}. }
    \resizebox{\linewidth}{!}{
        \begin{tabular}{ccccc}
            \toprule
                Dataset & \# Training Instance & \# Testing Instance & \# N-Shot for CoT & \# Total Instance \\
            \midrule
                MedQA-USMLE~\cite{jin2020disease} & 10178 & 2535 & 10 & 12723 \\
                MedQA-MCMLE~\cite{jin2020disease} & 27400 & 6840 & 11 & 34251 \\
                MedMCQA~\cite{pal2022medmcqa} & 182822 & 4170 & 13 & 187005 \\
                PubMedQA~\cite{jin2019pubmedqa} & - & 990 & 10 & 1000 \\
                MMLU-Med~\cite{hendryckstest2021} & - & 1080 & 9 & 1089 \\
                
                NEJMQA~\cite{katz2024gpt} & - & 655 & 10 & 655 \\
                MedXpertQA~\cite{zuo2025medxpertqa} & - & 2455 & 10 & 2455 \\
                RareArena~\cite{THUMedInfo_RareArena} & - & 72,661 & 10 & 72,661 \\
            \bottomrule
        \end{tabular}
    }
\end{table}

\subsection{Details of Baselines}
\label{appendix_subsec:baselines}


\begin{itemize}[leftmargin=*]
    \item \textbf{Frontier Models}: Frontier models are considered to represent the pinnacle of performance across various dimensions of LLMs, serving as the strongest baselines. Here, we have selected several state-of-the-art LLMs to establish the maximum potential of model performance on several medical benchmarks, including GPT-4o~\cite{hurst2024gpt}, Claude3.5-Sonnet~\cite{claude_anthropic_2024}, and DeepSeek-V3~\cite{liu2024deepseek}. 
    \item \textbf{Open-Sourced Medical Models}: These models stand for the domain-specific models that were specifically trained on medical data. 
    We have selected PMC-LLaMA-7B, PMC-LLaMA-13B~\cite{wu2024pmc}, MEDITRON-7B and MEDITRON-70B~\cite{chen2023meditron} to represent the open-sourced medical models for comparison to assess Med-R$^2$'s relative advantage compared to specialized medical AI systems. 
    \item \textbf{Direct Response}: We employ the base model to directly respond to medical queries without aid of additional training on any dataset or augmentation from external knowledge bases.
    \item \textbf{Vanilla RAG}~\cite{lewis2020retrieval}: It represents the most traditional and fundamental strategy for utilizing external knowledge bases to assist models in answering questions. In this study, we employ the same medical knowledge base as \ours, but directly utilizing the raw text retrieved by FAISS and combining it with the original query to test the model's performance on medical tasks. 
    \item \textbf{Within-Dataset Fine-Tuning}: The training and test sets originate from the same distribution, and models exhibit strong performance gains due to the homogeneity of the datasets. 
    \item \textbf{Cross-Dataset Fine-Tuning}: It poses a greater challenge by testing the model's ability to generalize across different distributions, demanding robust transfer learning capabilities, which are bolstered by the integration of rich external medical knowledge. 
    \item \textbf{LLM-AMT}~\cite{wang2024augmenting}: LLM-AMT is a RAG system specifically designed for clinical question answering. It incorporates common RAG components such as a query augmentor, textbook retriever, knowledge refiner, and an LLM reader. LLM-AMT leverages a collection of medical textbooks as an external indexable medical knowledge base. However, in this comparison, we aim to assess the RAG's capability to retrieve, filter, and apply evidence from the same external knowledge base. To control variables, we employ the LLM-AMT's process but replace the medical knowledge base with our own constructed medical retrieval corpus. 
\end{itemize}

\subsection{Hyper-Parameters Setting}
\label{appendix_subsec:hyper-parameters}


In the fine-grained reranking phase, we treat each factor as having equal importance, hence we set the weight controlling hyper-parameter $\alpha$ in \Cref{eq:fine_grained_reranking} to 1. 
For retrieval iteration settings, we performed iterative retrieval on each query within the datasets in \Cref{tab:datasets_stats}. 
Our analysis revealed that after approximately 5 iterations, the distribution of the retrieved document vectors stabilizes. Consequently, we set the maximum iterations $T$ to 5 in \Cref{alg:evidence_assessment}. 
After performing iterative retrieval for all samples in the dataset and calculating the average of the minimum distances in the retrieval space during iterations, we obtained a value of 6.85, which we adopted as our termination alteration threshold $\delta$ in \Cref{alg:evidence_assessment}. 



\begin{figure*}[th]
    \begin{center}
    \centerline{\includegraphics[width=\linewidth]{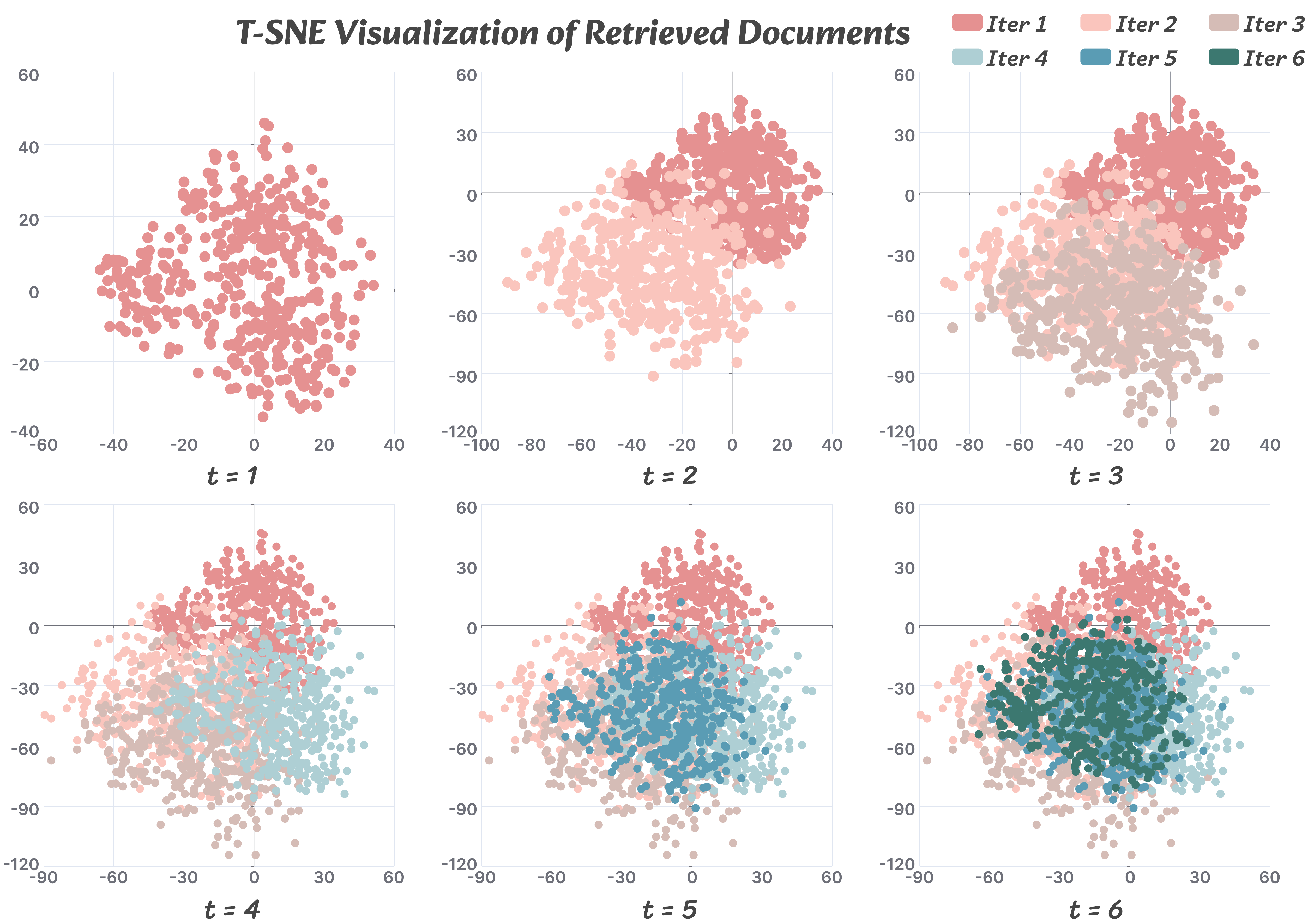}}
    \caption{A case of the t-SNE visualization for retrieved documents. 
    We visualized the projections of document embeddings onto a 2-D plane across different iterations. It is evident that aside from the significant variation in the retrieval document vector space at $t = 2$, the semantics of the retrieved documents tend to stabilize in subsequent iterations. It occurs because we did not incorporate the model's CoT sequence for the initial retrieval round ($t = 1$). Instead, we began to include it starting from the second iteration onwards ($t = 2$) to avoid the generation of sequences that are not only unrealistic but also detrimental to the precision of evidence document retrieval.}
    \label{fig:tSNE_iteration}
    \end{center}
\end{figure*}

\section{Scaling Analysis and Ablation Details}
\label{sec:appendix_ablation}

In this section, we conduct a more detailed analysis of the impact of context window size and model parameter scale on our Med-R$^2$. 


\begin{table*}[ht]
    \centering
    \caption{\label{tab:context_window_scaling} 
    Scaling analysis of context window and model size for \ours.
}
    \resizebox{0.85\textwidth}{!}{
        \begin{tabular}{cccccccc}
            \toprule
                Model & \multicolumn{1}{|c}{Method} & \multicolumn{1}{|c}{MedQA-USMLE} & MedQA-MCMLE & MedMCQA &  \multicolumn{1}{|c}{PubMedQA} &  \multicolumn{1}{c|}{MMLU-Med} & Average \\
            \midrule
            \multicolumn{8}{c}{\cellcolor[HTML]{EFEFEF}\textit{\textbf{4K}}} \\
            \midrule
            \multirow{2}{*}{LLaMA3.1-8B} & \multicolumn{1}{|c}{Direct Response} & \multicolumn{1}{|c}{31.16} & 41.45 & 30.02 & \multicolumn{1}{|c}{36.17} & 37.12 & \multicolumn{1}{|c}{\cellcolor[HTML]{FFFBCF}35.18} \\
            & \ours & \multicolumn{1}{|c}{77.01} & 84.16 & 52.33 & \multicolumn{1}{|c}{52.75} & 45.19 & \multicolumn{1}{|c}{\cellcolor[HTML]{FFFBCF}62.29} \\
            \midrule
            \multirow{2}{*}{Qwen2.5-14B} & \multicolumn{1}{|c}{Direct Response} & \multicolumn{1}{|c}{50.01} & 65.23 & 42.85 & \multicolumn{1}{|c}{56.93} & 71.60 & \multicolumn{1}{|c}{\cellcolor[HTML]{FFFBCF}57.32} \\
            & \ours & \multicolumn{1}{|c}{54.34} & 80.19 & 48.36 & \multicolumn{1}{|c}{68.32} & 84.03 & \multicolumn{1}{|c}{\cellcolor[HTML]{FFFBCF}67.05} \\
            \midrule
            \multirow{2}{*}{Qwen2.5-32B} & \multicolumn{1}{|c}{Direct Response} & \multicolumn{1}{|c}{16.23} & 87.07 & 66.44 & \multicolumn{1}{|c}{68.66} & 80.19 & \multicolumn{1}{|c}{\cellcolor[HTML]{FFFBCF}63.72} \\
            & \ours & \multicolumn{1}{|c}{24.43} & 90.01 & 69.09 & \multicolumn{1}{|c}{68.36} & 84.95 & \multicolumn{1}{|c}{\cellcolor[HTML]{FFFBCF}67.37} \\
            \midrule
            \multirow{2}{*}{LLaMA3.1-70B} & \multicolumn{1}{|c}{Direct Response} & \multicolumn{1}{|c}{46.43} & 58.36 & 62.33 & \multicolumn{1}{|c}{66.81} & 71.33 & \multicolumn{1}{|c}{\cellcolor[HTML]{FFFBCF}61.05} \\
            & \ours & \multicolumn{1}{|c}{86.37} & 84.58 & 73.36 & \multicolumn{1}{|c}{78.24} & 82.82 & \multicolumn{1}{|c}{\cellcolor[HTML]{FFFBCF}81.07} \\
            \midrule
            \multicolumn{8}{c}{\cellcolor[HTML]{EFEFEF}\textit{\textbf{8K}}} \\
            \midrule
            \multirow{2}{*}{LLaMA3.1-8B} & \multicolumn{1}{|c}{Direct Response} & \multicolumn{1}{|c}{31.34} & 41.64 & 29.54 & \multicolumn{1}{|c}{36.71} & 38.65 & \multicolumn{1}{|c}{\cellcolor[HTML]{FFFBCF}35.58} \\
            & \multicolumn{1}{|c}{\ours} & \multicolumn{1}{|c}{73.16} & 82.61 & 47.72 & \multicolumn{1}{|c}{58.72} & 48.74 & \multicolumn{1}{|c}{\cellcolor[HTML]{FFFBCF}62.19} \\
            \midrule
            \multirow{2}{*}{Qwen2.5-14B} & \multicolumn{1}{|c}{Direct Response} & \multicolumn{1}{|c}{50.21} & 64.21 & 41.34 & \multicolumn{1}{|c}{57.18} & 70.28 & \multicolumn{1}{|c}{\cellcolor[HTML]{FFFBCF}56.64} \\
            & \multicolumn{1}{|c}{\ours}  & \multicolumn{1}{|c}{58.50} & 86.86 & 53.40 & \multicolumn{1}{|c}{69.74} & 84.29 & \multicolumn{1}{|c}{\cellcolor[HTML]{FFFBCF}70.56} \\
            \midrule
            \multirow{2}{*}{Qwen2.5-32B} & \multicolumn{1}{|c}{Direct Response} & \multicolumn{1}{|c}{15.75} & 86.82 & 66.39 & \multicolumn{1}{|c}{68.64} & 79.73 & \multicolumn{1}{|c}{\cellcolor[HTML]{FFFBCF}63.51} \\
            & \multicolumn{1}{|c}{\ours} & \multicolumn{1}{|c}{25.28} & 89.36 & 75.55 & \multicolumn{1}{|c}{68.87} & 84.21 & \multicolumn{1}{|c}{\cellcolor[HTML]{FFFBCF}68.65} \\
            \midrule
            \multirow{2}{*}{LLaMA3.1-70B} & \multicolumn{1}{|c}{Direct Response} & \multicolumn{1}{|c}{47.99} & 57.78 & 61.69 & \multicolumn{1}{|c}{67.20} & 70.69 & \multicolumn{1}{|c}{\cellcolor[HTML]{FFFBCF}61.07} \\
            & \multicolumn{1}{|c}{\ours} & \multicolumn{1}{|c}{85.72} & 86.80 & 78.21 & \multicolumn{1}{|c}{78.66} & 84.82 & \multicolumn{1}{|c}{\cellcolor[HTML]{FFFBCF}82.84} \\
            \midrule
            \multicolumn{8}{c}{\cellcolor[HTML]{EFEFEF}\textit{\textbf{16K}}} \\
            \midrule
            \multirow{2}{*}{LLaMA3.1-8B} & \multicolumn{1}{|c}{Direct Response} & \multicolumn{1}{|c}{31.39} & 41.43 & 29.72 & \multicolumn{1}{|c}{36.70} & 37.78 & \multicolumn{1}{|c}{\cellcolor[HTML]{FFFBCF}35.40} \\
            & \multicolumn{1}{|c}{\ours} & \multicolumn{1}{|c}{72.54} & 77.22 & 45.13 & \multicolumn{1}{|c}{57.45} & 46.75 & \multicolumn{1}{|c}{\cellcolor[HTML]{FFFBCF}59.82} \\
            \midrule
            \multirow{2}{*}{Qwen2.5-14B} & \multicolumn{1}{|c}{Direct Response} & \multicolumn{1}{|c}{49.69} & 66.44 & 41.84 & \multicolumn{1}{|c}{57.07} & 69.10 & \multicolumn{1}{|c}{\cellcolor[HTML]{FFFBCF}56.83} \\
            & \multicolumn{1}{|c}{\ours}  & \multicolumn{1}{|c}{55.46} & 83.01 & 52.67 & \multicolumn{1}{|c}{67.23} & 83.68 & \multicolumn{1}{|c}{\cellcolor[HTML]{FFFBCF}68.41} \\
            \midrule
            \multirow{2}{*}{Qwen2.5-32B} & \multicolumn{1}{|c}{Direct Response} & \multicolumn{1}{|c}{15.46} & 86.83 & 66.52 & \multicolumn{1}{|c}{68.48} & 80.07 & \multicolumn{1}{|c}{\cellcolor[HTML]{FFFBCF}63.47} \\
            & \multicolumn{1}{|c}{\ours} & \multicolumn{1}{|c}{28.02} & 89.32 & 70.79 & \multicolumn{1}{|c}{70.51} & 85.72 & \multicolumn{1}{|c}{\cellcolor[HTML]{FFFBCF}68.87} \\
            \midrule
            \multirow{2}{*}{LLaMA3.1-70B} & \multicolumn{1}{|c}{Direct Response} & \multicolumn{1}{|c}{47.46} & 57.61 & 61.62 & \multicolumn{1}{|c}{67.45} & 69.09 & \multicolumn{1}{|c}{\cellcolor[HTML]{FFFBCF}60.65} \\
            & \multicolumn{1}{|c}{\ours} & \multicolumn{1}{|c}{86.43} & 88.15 & 74.64 & \multicolumn{1}{|c}{80.94} & 87.24 & \multicolumn{1}{|c}{\cellcolor[HTML]{FFFBCF}83.48} \\
            \midrule
            \multicolumn{8}{c}{\cellcolor[HTML]{EFEFEF}\textit{\textbf{32K}}} \\
            \midrule
            \multirow{2}{*}{LLaMA3.1-8B} & \multicolumn{1}{|c}{Direct Response} & \multicolumn{1}{|c}{31.25} & 40.51 & 30.58 & \multicolumn{1}{|c}{36.62} & 38.44 & \multicolumn{1}{|c}{\cellcolor[HTML]{FFFBCF}35.48} \\
            & \multicolumn{1}{|c}{\ours} & \multicolumn{1}{|c}{62.86} & 68.67 & 41.65 & \multicolumn{1}{|c}{57.22} & 43.44 & \multicolumn{1}{|c}{\cellcolor[HTML]{FFFBCF}54.77} \\
            \midrule
            \multirow{2}{*}{Qwen2.5-14B} & \multicolumn{1}{|c}{Direct Response} & \multicolumn{1}{|c}{50.40} & 66.31 & 41.27 & \multicolumn{1}{|c}{58.24} & 69.31 & \multicolumn{1}{|c}{\cellcolor[HTML]{FFFBCF}57.11} \\
            & \multicolumn{1}{|c}{\ours}  & \multicolumn{1}{|c}{54.78} & 78.41 & 50.76 & \multicolumn{1}{|c}{65.73} & 83.23 & \multicolumn{1}{|c}{\cellcolor[HTML]{FFFBCF}66.58} \\
            \midrule
            \multirow{2}{*}{Qwen2.5-32B} & \multicolumn{1}{|c}{Direct Response} & \multicolumn{1}{|c}{15.61} & 86.85 & 64.98 & \multicolumn{1}{|c}{69.16} & 80.10 & \multicolumn{1}{|c}{\cellcolor[HTML]{FFFBCF}63.34} \\
            & \multicolumn{1}{|c}{\ours} & \multicolumn{1}{|c}{27.05} & 89.29 & 70.80 & \multicolumn{1}{|c}{71.07} & 84.56 & \multicolumn{1}{|c}{\cellcolor[HTML]{FFFBCF}68.55} \\
            \midrule
            \multirow{2}{*}{LLaMA3.1-70B} & \multicolumn{1}{|c}{Direct Response} & \multicolumn{1}{|c}{46.31} & 57.89 & 61.87 & \multicolumn{1}{|c}{67.32} & 69.96 & \multicolumn{1}{|c}{\cellcolor[HTML]{FFFBCF}60.67} \\
            & \multicolumn{1}{|c}{\ours} & \multicolumn{1}{|c}{83.36} & 85.58 & 75.93 & \multicolumn{1}{|c}{78.22} & 86.61 & \multicolumn{1}{|c}{\cellcolor[HTML]{FFFBCF}81.94} \\
            \midrule
            \multicolumn{8}{c}{\cellcolor[HTML]{EFEFEF}\textit{\textbf{64K}}} \\
            \midrule
            \multirow{2}{*}{LLaMA3.1-8B} & \multicolumn{1}{|c}{Direct Response} & \multicolumn{1}{|c}{31.92} & 39.85 & 30.24 & \multicolumn{1}{|c}{36.68} & 37.52 & \multicolumn{1}{|c}{\cellcolor[HTML]{FFFBCF}35.24} \\
            & \multicolumn{1}{|c}{\ours} & \multicolumn{1}{|c}{62.57} & 62.51 & 40.97 & \multicolumn{1}{|c}{52.44} & 45.73 & \multicolumn{1}{|c}{\cellcolor[HTML]{FFFBCF}52.84} \\
            \midrule
            \multirow{2}{*}{Qwen2.5-14B} & \multicolumn{1}{|c}{Direct Response} & \multicolumn{1}{|c}{49.88} & 64.76 & 42.02 & \multicolumn{1}{|c}{57.06} & 69.42 & \multicolumn{1}{|c}{\cellcolor[HTML]{FFFBCF}56.63} \\
            & \multicolumn{1}{|c}{\ours}  & \multicolumn{1}{|c}{54.84} & 78.75 & 50.64 & \multicolumn{1}{|c}{63.06} & 80.51 & \multicolumn{1}{|c}{\cellcolor[HTML]{FFFBCF}65.56} \\
            \midrule
            \multirow{2}{*}{Qwen2.5-32B} & \multicolumn{1}{|c}{Direct Response} & \multicolumn{1}{|c}{15.89} & 86.81 & 65.12 & \multicolumn{1}{|c}{68.65} & 79.29 & \multicolumn{1}{|c}{\cellcolor[HTML]{FFFBCF}63.15} \\
            & \multicolumn{1}{|c}{\ours} & \multicolumn{1}{|c}{27.87} & 88.75 & 70.42 & \multicolumn{1}{|c}{70.65} & 84.54 & \multicolumn{1}{|c}{\cellcolor[HTML]{FFFBCF}68.45} \\
            \midrule
            \multirow{2}{*}{LLaMA3.1-70B} & \multicolumn{1}{|c}{Direct Response} & \multicolumn{1}{|c}{47.17} & 57.43 & 61.79 & \multicolumn{1}{|c}{67.18} & 69.37 & \multicolumn{1}{|c}{\cellcolor[HTML]{FFFBCF}60.59} \\
            & \multicolumn{1}{|c}{\ours} & \multicolumn{1}{|c}{83.55} & 84.86 & 74.09 & \multicolumn{1}{|c}{78.79} & 85.70 & \multicolumn{1}{|c}{\cellcolor[HTML]{FFFBCF}81.40} \\
            \midrule
            \multicolumn{8}{c}{\cellcolor[HTML]{EFEFEF}\textit{\textbf{128K}}} \\
            \midrule
            \multirow{2}{*}{LLaMA3.1-8B} & \multicolumn{1}{|c}{Direct Response} & \multicolumn{1}{|c}{31.14} & 39.97 & 30.69 & \multicolumn{1}{|c}{36.43} & 38.26 & \multicolumn{1}{|c}{\cellcolor[HTML]{FFFBCF}35.30} \\
            & \multicolumn{1}{|c}{\ours} & \multicolumn{1}{|c}{56.06} & 59.58 & 32.48 & \multicolumn{1}{|c}{47.64} & 44.82 & \multicolumn{1}{|c}{\cellcolor[HTML]{FFFBCF}48.12} \\
            \midrule
            \multirow{2}{*}{Qwen2.5-14B} & \multicolumn{1}{|c}{Direct Response} & \multicolumn{1}{|c}{51.38} & 65.04 & 40.73 & \multicolumn{1}{|c}{57.50} & 70.90 & \multicolumn{1}{|c}{\cellcolor[HTML]{FFFBCF}57.11} \\
            & \multicolumn{1}{|c}{\ours}  & \multicolumn{1}{|c}{52.79} & 75.53 & 49.52 & \multicolumn{1}{|c}{63.96} & 77.61 & \multicolumn{1}{|c}{\cellcolor[HTML]{FFFBCF}63.88} \\
            \midrule
            \multirow{2}{*}{Qwen2.5-32B} & \multicolumn{1}{|c}{Direct Response} & \multicolumn{1}{|c}{16.11} & 85.77 & 66.29 & \multicolumn{1}{|c}{68.12} & 79.84 & \multicolumn{1}{|c}{\cellcolor[HTML]{FFFBCF}63.23} \\
            & \multicolumn{1}{|c}{\ours} & \multicolumn{1}{|c}{27.09} & 88.21 & 69.17 & \multicolumn{1}{|c}{70.99} & 83.87 & \multicolumn{1}{|c}{\cellcolor[HTML]{FFFBCF}67.87} \\
            \midrule
            \multirow{2}{*}{LLaMA3.1-70B} & \multicolumn{1}{|c}{Direct Response} & \multicolumn{1}{|c}{47.26} & 57.37 & 63.28 & \multicolumn{1}{|c}{68.12} & 70.03 & \multicolumn{1}{|c}{\cellcolor[HTML]{FFFBCF}61.21} \\
            & \multicolumn{1}{|c}{\ours} & \multicolumn{1}{|c}{83.84} & 84.07 & 74.97 & \multicolumn{1}{|c}{77.87} & 84.93 & \multicolumn{1}{|c}{\cellcolor[HTML]{FFFBCF}81.14} \\
            \bottomrule
        \end{tabular}
    }
\end{table*}

\textbf{Scaling for Context Window} \quad 
The training context windows length for 
the LLaMA3.1 series models are explicitly designed to extend up to a context length of 128K. The Qwen2.5 series, trained at a 32K context length, can also be extended to 128K by modifying the \texttt{max\_position\_embedding} in the \texttt{config.json} file through \texttt{Yarn}. Consequently, we have chosen two models each from the LLaMA3.1 series and the Qwen2.5 series to conduct inference tests on context window length expansion. 
Experimental results are outlined in \Cref{tab:context_window_scaling} and \Cref{fig:context_window_scaling}.



\section{Case Studies}
\label{sec:appendix_case_study}

\begin{figure*}[t]
    \begin{center}
    \centerline{\includegraphics[width=\linewidth]{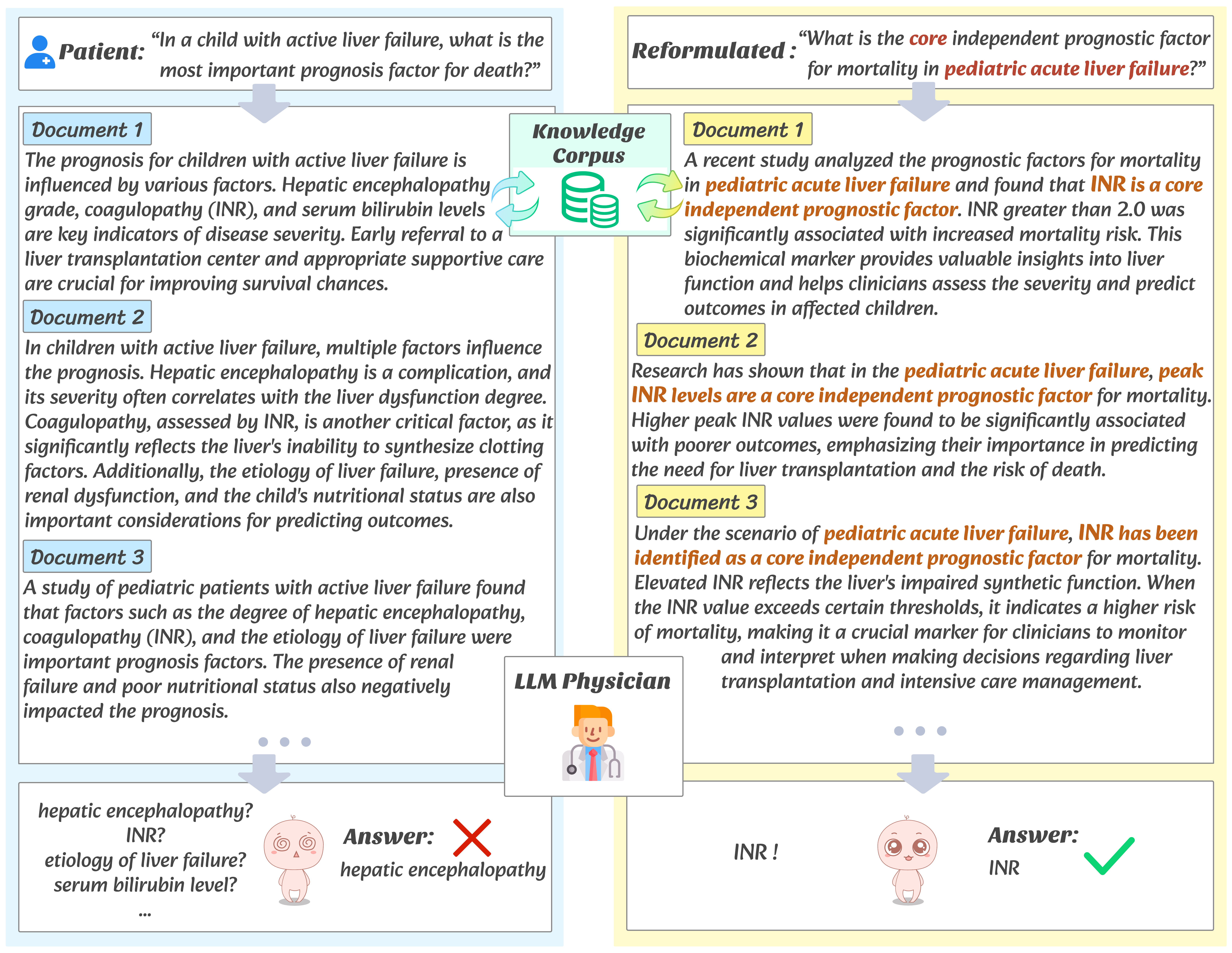}}
    \caption{\label{fig:case_query} 
    A case study of \textbf{\textit{query reformulation}}. 
    We \textcolor{BF6119}{\textbf{\textit{highlight}}} the sentences within the retrieved documents that exhibit a high degree of relevance to the keywords in the query and the final answer after the query reformulation process. It is evident that prior to query reformulation \textbf{(left)}, the retrieved evidence contains descriptions of acute liver failure mortality prognoses that are dispersed, including ``hepatic encephalopathy'', ``INR'', ``serum bilirubin level'', etc., without emphasizing the most critical prognostic factors. Consequently, the model is unable to extract the correct answers when answering medical questions based on such documents. In contrast, after the query reformulation process \textbf{(right)}, it can be observed that the retrieved documents consistently identify ``INR'' as a core prognostic feature, providing effective evidential support for the model's responses to medical inquiries.}
    \end{center}
\end{figure*}

\begin{figure*}[ht]
    \vskip 0.15in
    \begin{center}
    \centerline{\includegraphics[width=\linewidth]{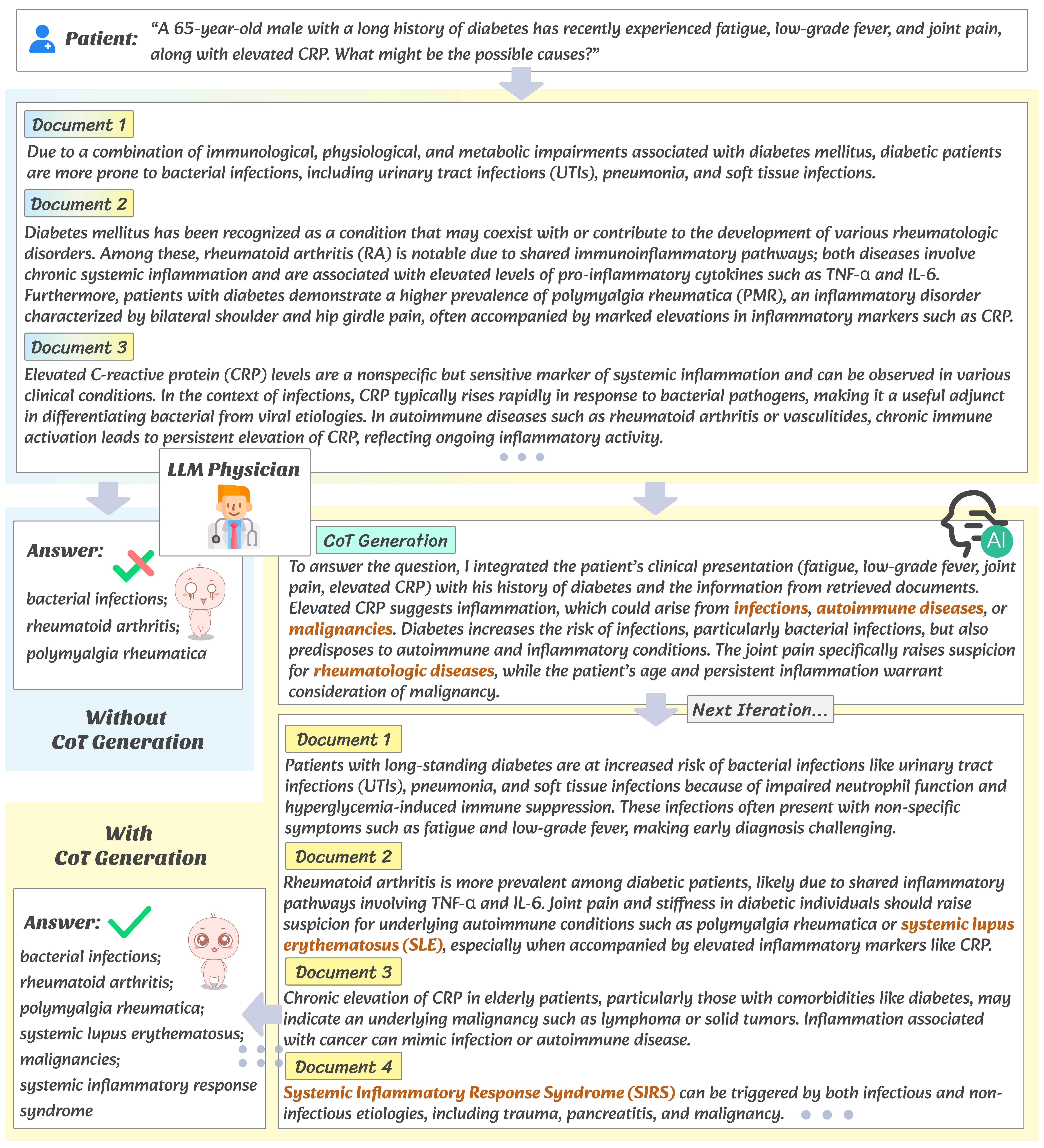}}
    \caption{\label{fig:case_cot} 
    A case study of \textbf{\textit{CoT generation}}. 
    We \textcolor{BF6119}{\textbf{\textit{highlight}}} the keywords within the generated CoT sequences and retrieved documents that play a pivotal role in both retrieval recall and answer inference. It is observed that when the model generates CoT sequences based on the initial set of retrieved documents, it incorporates keywords such as ``infections'', ``autoimmune diseases'', ``malignancies'', and ``rheumatologic diseases''. The inclusion of these keywords enhances the comprehensiveness of the evidence documents retrieved in subsequent iterations. For instance, the second round of document retrieval encompasses causes like ``systemic lupus erythematosus (SLE)'' and ``systemic inflammatory response syndrome (SIRS)''. Through such iterative process, the model progressively accumulates more relevant information, leading to a final answer that is notably more accurate and comprehensive compared to approaches that do \textit{not} incorporate the CoT generation module.}
    \end{center}
    \vskip -0.2in
\end{figure*}

Here, we provide concrete examples of \textit{query reformulation} and \textit{CoT generation} to demonstrate the critical importance of these components. 
(1) \textbf{Query Reformulation}: As depicted in \Cref{fig:case_query}, the example utilizes the standard medical term ``acute liver failure'' with the specification ``pediatric'' to focus on the child population, which facilitates retrieval precision. It is more suitable for scenarios that require rapid positioning of medical concepts. Moreover, the term ``core'' is more professional and precise than ``the most important'', offering a distinct advantage in evidence retrieval. 
(2) \textbf{CoT Generation}: As illustrated in \Cref{fig:case_cot}, its advantage lies in further optimizing the precision of retrieval based on the content from the previous round of the CoT process. During the CoT process, the addition of keywords such as ``infections'', ``autoimmune diseases'', ``malignancies'', and ``rheumatologic diseases'' improved the recall accuracy for potential causes in the next iteration. 


\clearpage

\clearpage


\clearpage

\section{Prompts}
\label{sec:appendix_prompts}

We present the prompts employed throughout our pipeline in \ours. 
Specially, prompts for query reformulation according to the Evidence-Based Medicine (EBM) categories can be found in \Cref{tab:EBM_category}.

\begin{tcolorbox}[breakable, title=\textbf{Prompt: Evidence-Based Medicine (EBM) Category Classification}, colback=gray!10, colframe=gray!50!black, boxrule=1pt]
You are an expert in sentence annotation within the medical field. There are 6 categories of clinical questions: Prognosis, Therapy, Etiology, Diagnosis, Prevention, and Cost. Please classify the following text fragment based on their purpose and structure by providing only the category name without additional commentary:

\vspace{0.3cm}

\textbf{\# Question}\\
\{question\}

\end{tcolorbox}

\begin{tcolorbox}[breakable, title=\textbf{Prompt: General Natural Language Question Category Classification}, colback=gray!10, colframe=gray!50!black, boxrule=1pt]
You are an expert in natural language question annotation. Given the following 12 categories of question types: Factual, Definitional, Explanatory, Descriptive, Directive, Opinion, Comparative, Evaluative, Hypothetical, Procedural, Referential, and Verification.  Please classify the following text fragment based on their purpose and structure by providing only the category name without additional commentary:

\vspace{0.3cm}

\textbf{\# Question}\\
\{question\}

\end{tcolorbox}

\begin{tcolorbox}[breakable, title=\textbf{Prompt: Retrieved Document Category Classification}, colback=gray!10, colframe=gray!50!black, boxrule=1pt]
You are an expert in sentence annotation within the medical field. There are 16 categories of documents: Argumentation, Definition, Description, Explanation, Purpose, Narration, Process, Instruction, Command, Problem-Solving, Comparison, Evaluation, Classification, Condition, Prediction, Cause-and-Effect. Please classify the following text fragment based on their purpose and structure by providing the probability distribution of its belonging to each category, where the sum of probabilities across all categories equals 1, without additional commentary:

\vspace{0.3cm}

\textbf{\# Document}\\
\{retrieved\_document\}

\vspace{0.1cm}

\hdashrule{\textwidth}{1pt}{3pt}

\vspace{0.1cm}

\textbf{Output Format:}  
\begin{verbatim}
```json
{
    "Argumentation": "",
    "Definition": "",
    ...
}
'''
\end{verbatim}

\end{tcolorbox}

\begin{tcolorbox}[breakable, title=\textbf{Prompt: Hierarchy of Evidence Judgement}, colback=gray!10, colframe=gray!50!black, boxrule=1pt]
You are an expert in evidence quality annotation within the medical field. There are 9 quality levels of evidence, ranging from the highest to the lowest as follows: Meta-Analyses, Systematic Reviews, Evidence-Based Practice Guidelines, Randomized Controlled Trials, Non-Randomized Controlled Trials, Cohort Studies, Case Series or Studies, Individual Case Reports, Expert Opinion. Please classify the following evidence document based on its structure and characteristics, providing only the names of the levels, without any additional description:

\vspace{0.3cm}

\textbf{\# Evidence}\\
\{retrieved\_document\}

\end{tcolorbox}

\begin{tcolorbox}[breakable, title=\textbf{Prompt: Chain-of-Thought (CoT) Generator}, colback=gray!10, colframe=gray!50!black, boxrule=1pt]
Given the provided \texttt{[Context]}, \texttt{[Question]}, as well as the \texttt{[Retrieved Documents]}, please provide an answer that includes your thought process. Specifically:

\begin{enumerate}
    \item \textbf{\textit{\underline{Analyze the Question}}}: Carefully analyze the \texttt{[Question]} to understand what information is being sought.
    \item \textbf{\textit{\underline{Review Provided Context}}}: Examine the \texttt{[Context]} for any background information that can help frame the answer.
    \item \textbf{\textit{\underline{Consult Retrieved Documents}}}: Go through the snippets of \texttt{[Retrieved Documents]} to identify sections that are directly related to \texttt{[Question]}.
    \item \textbf{\textit{\underline{Identify Key Information}}}: Highlight the key points from the \texttt{[Retrieved Documents]} that address the question's requirements.
    \item \textbf{\textit{\underline{Construct Thought Process}}}: Explain how you used the information from the \texttt{[Context]} and the retrieved documents to form your understanding and construct your answer.
    \item \textbf{\textit{\underline{Provide Answer}}}: Finally, give a clear and concise answer to the \texttt{[Question]}, supported by the analysis of the \texttt{[Retrieved Documents]}.
\end{enumerate}

Please present your response in a way that clearly shows your reasoning and the sources of information you relied on.

\vspace{0.3cm}

\textbf{\# Context}\\
\{context\} \textcolor[HTML]{0000FF}{[optional]}

\vspace{0.3cm}

\textbf{\# Question}\\
\{question\}

\vspace{0.3cm}

\textbf{\# Retrieved Documents}\\
\{retrieved\_documents\} \textcolor[HTML]{0000FF}{[optional]}

\end{tcolorbox}




\end{document}